\documentclass[10pt,twocolumn,letterpaper]{article}

\usepackage[pagenumbers]{cvpr} 

\usepackage[dvipsnames]{xcolor}

\usepackage{times}
\usepackage{epsfig}
\usepackage{graphicx}
\usepackage{amsmath}
\usepackage{amssymb}

\usepackage{booktabs}
\usepackage{multirow}
\usepackage{bm}
\usepackage{tikz}
\usepackage{comment}
\usepackage{color}
\usepackage{bbding}
\usepackage{gensymb}
\usepackage{colortbl}
\usepackage{algorithmicx,algorithm}
\usepackage[noend]{algpseudocode}
\usepackage[accsupp]{axessibility} 

\definecolor{cvprblue}{rgb}{0.21,0.49,0.74}
\usepackage[pagebackref,breaklinks,colorlinks,citecolor=cvprblue]{hyperref}

\newcommand\extrafootertext[1]{%
    \bgroup
    \renewcommand\thefootnote{\fnsymbol{footnote}}%
    \renewcommand\thempfootnote{\fnsymbol{mpfootnote}}%
    \footnotetext[0]{#1}%
    \egroup
}

\title{SAM-6D: Segment Anything Model Meets Zero-Shot 6D Object Pose Estimation}

\author{
   Jiehong Lin$^{1,2}$\footnotemark \and Lihua Liu$^{3 *}$ \and Dekun Lu$^3$ \and Kui Jia$^2$\footnotemark \and
   $^1$DexForce Co. Ltd., Shenzhen \\ $^2$School of Data Science, The Chinese University of Hong Kong, Shenzhen  \\
   $^3$South China University of Technology, Guangzhou   \\
   {\tt\small \url{https://github.com/JiehongLin/SAM-6D}}
}

\begin{document}

\twocolumn[{
\maketitle\centering
\captionsetup{type=figure}
\vspace{-0.75cm}
\includegraphics[width=0.99\textwidth]{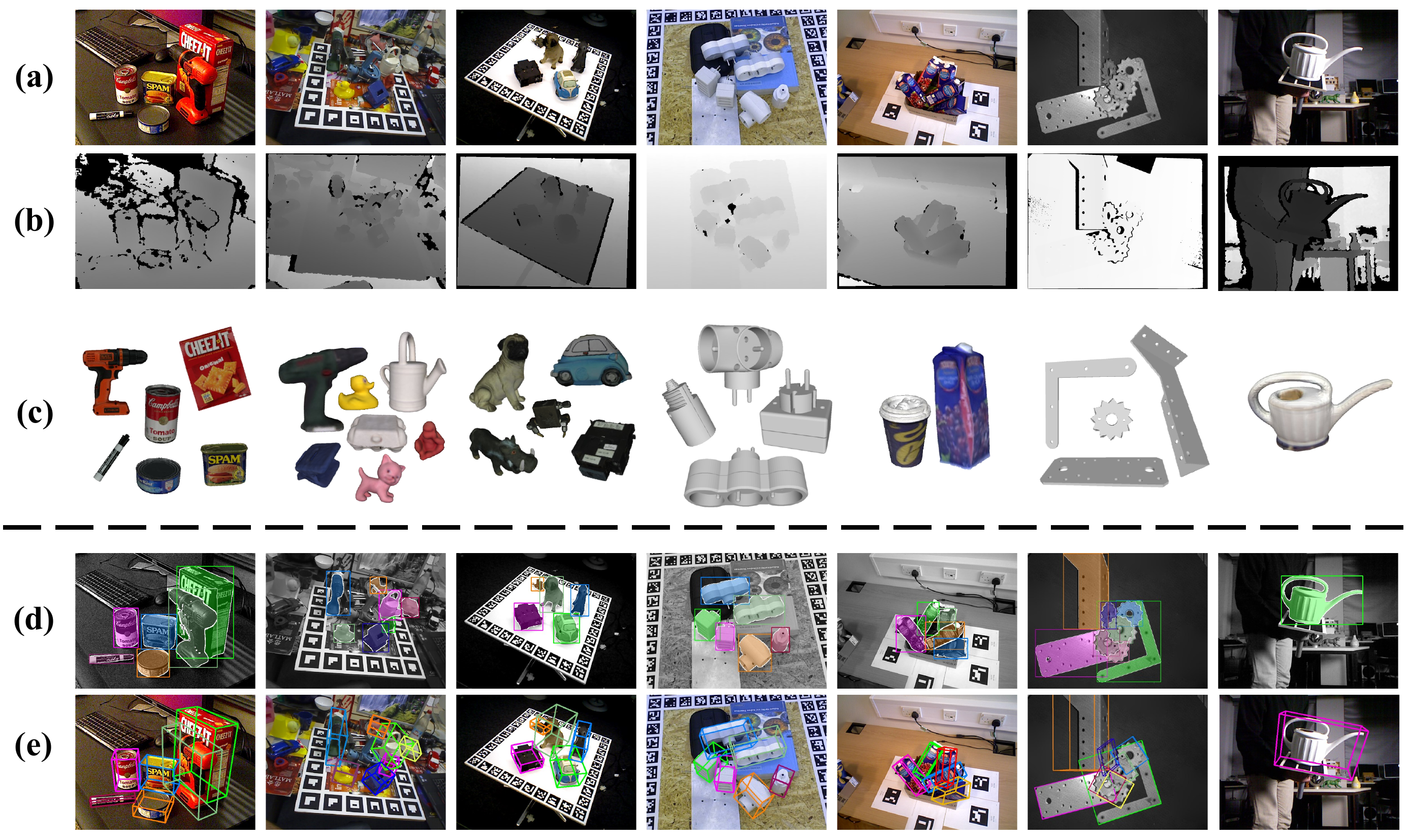}\vspace{-4mm}
\captionof{figure}{We present \textbf{SAM-6D} for zero-shot 6D object pose estimation. SAM-6D takes an RGB image (a) and a depth map (b) of a cluttered scene as inputs, and performs instance segmentation (d) and pose estimation (e) for novel objects (c). We present the qualitative results of SAM-6D on the seven core datasets of the BOP benchmark \cite{BOP}, including YCB-V, LM-O, HB, T-LESS, IC-BIN, ITODD and TUD-L, arranged from left to right. Best view in the electronic version.}
\label{fig:vis}\vspace{4mm}
}]

\extrafootertext{
{{{\textsuperscript{*} Equal contribution.} \textsuperscript{\dag} Corresponding author $<$kuijia@gmail.com$>$.}}
}

\begin{abstract}

Zero-shot 6D object pose estimation involves the detection of novel objects with their 6D poses in cluttered scenes, presenting significant challenges for model generalizability. Fortunately, the recent Segment Anything Model (SAM) has showcased remarkable zero-shot transfer performance, which provides a promising solution to tackle this task. Motivated by this, we introduce SAM-6D, a novel framework designed to realize the task through two steps, including instance segmentation and pose estimation. Given the target objects, SAM-6D employs two dedicated sub-networks, namely Instance Segmentation Model (ISM) and Pose Estimation Model (PEM), to perform these steps on cluttered RGB-D images. ISM takes SAM as an advanced starting point to generate all possible object proposals and selectively preserves valid ones through meticulously crafted object matching scores in terms of semantics, appearance and geometry. By treating pose estimation as a partial-to-partial point matching problem, PEM performs a two-stage point matching process featuring a novel design of background tokens to construct dense 3D-3D correspondence, ultimately yielding the pose estimates. Without bells and whistles, SAM-6D outperforms the existing methods on the seven core datasets of the BOP Benchmark for both instance segmentation and pose estimation of novel objects. 
\end{abstract}    

\section{Introduction}
\label{sec:intro}

Object pose estimation is fundamental in many real-world applications, such as robotic manipulation and augmented reality. Its evolution has been significantly influenced by the emergence of deep learning models. The most studied task in this field is \textit{Instance-level 6D Pose Estimation}  \cite{xiang2017posecnn, wang2019densefusion, wang2021gdr, su2022zebrapose, he2020pvn3d, he2021ffb6d}, which demands annotated training images of the target objects, thereby making the deep models object-specific. Recently, the research emphasis gradually shifts towards the task of \textit{Category-level 6D Pose Estimation} \cite{wang2019normalized, spd, lin2021dualposenet, lin2021sparse, lin2022category, chen2021sgpa, lin2023vi} for handling unseen objects, yet provided they belong to certain categories of interest. In this paper, we thus delve into a broader task setting of \textit{Zero-shot 6D Object Pose Estimation} \cite{MegaPose, ZeroPose}, which aspires to detect all instances of novel objects, unseen during training, and estimate their 6D poses. Despite its significance, this zero-shot setting presents considerable challenges in both object detection and pose estimation.

\begin{figure}[t]
  \centering
   \includegraphics[width=1.0\linewidth]{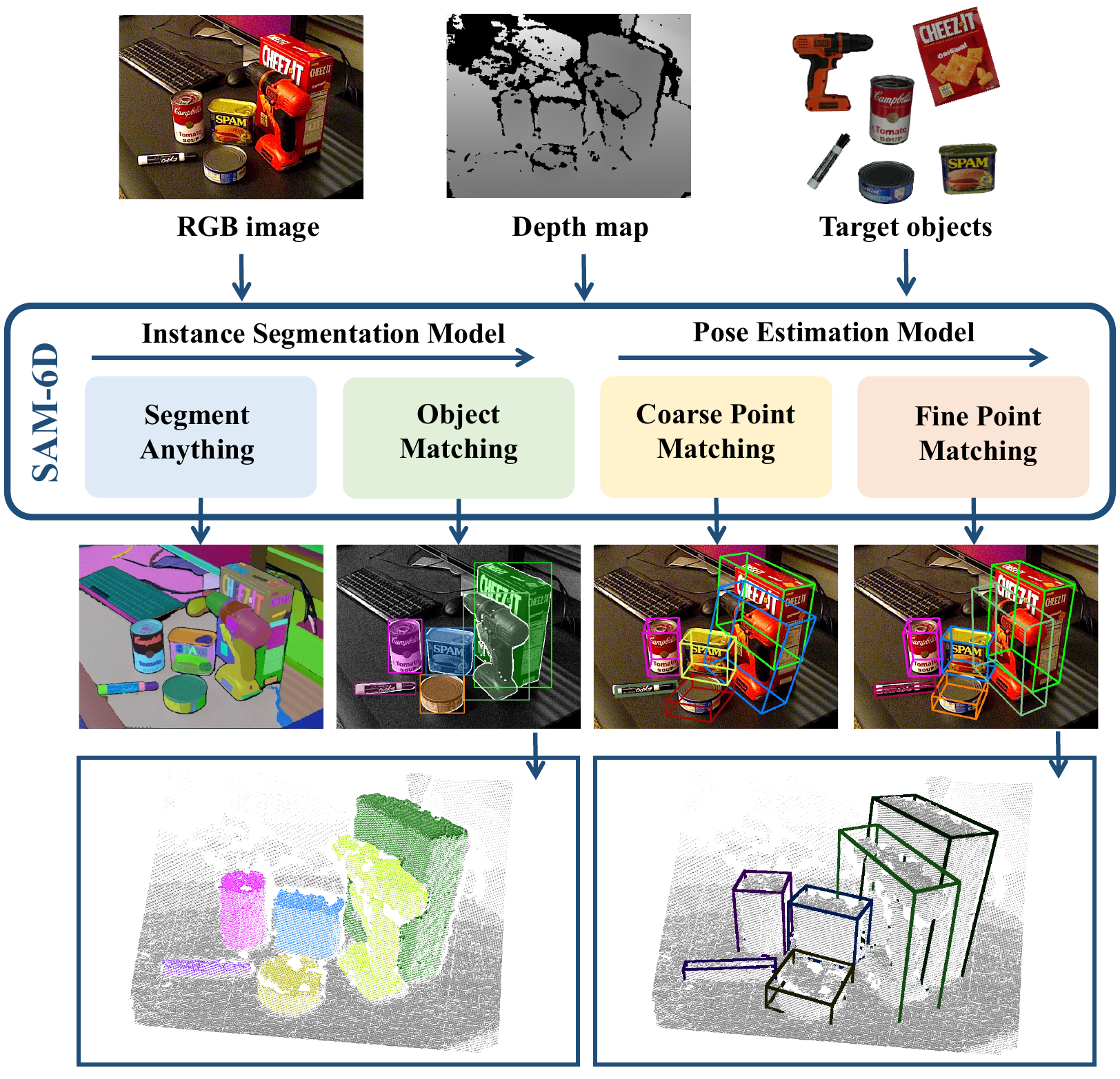}
   \vspace{-0.7cm}
   \caption{An overview of our proposed SAM-6D, which consists of an Instance Segmentation Model (ISM) and a Pose Estimation Model (PEM) for joint instance segmentation and pose estimation of novel objects in RGB-D images. ISM leverages the Segment Anything Model (SAM) \cite{SAM} to generate all possible proposals and selectively retains valid ones based on object matching scores. PEM involves two stages of point matching, from coarse to fine, to establish 3D-3D correspondence and calculate object poses for all valid proposals. Best view in the electronic version.}
   \vspace{-0.3cm}
   \label{fig:head}
\end{figure}

Recently, Segment Anything Model (SAM) \cite{SAM} has garnered attention due to its remarkable zero-shot segmentation performance, which enables prompt segmentation with a variety of prompts, \eg,  points, boxes, texts or masks. By prompting SAM with evenly sampled 2D grid points, one can generate potential class-agnostic object proposals, which may be highly beneficial for zero-shot 6D object pose estimation. To this end, we propose a novel framework, named \textbf{SAM-6D}, which employs SAM as an advanced starting point for the focused zero-shot task. Fig. \ref{fig:head} gives an overview illustration of SAM-6D. Specifically, SAM-6D employs an \textbf{Instance Segmentation Model} (ISM) to realize instance segmentation of novel objects by enhancing SAM with a carefully crafted object matching score, and a \textbf{Pose Estimation Model} (PEM) to solve object poses through a two-stage process of partial-to-partial point matching.

The Instance Segmentation Model (ISM) is developed using SAM to take advantage of its zero-shot abilities for generating all possible class-agnostic proposals, and then assigns a meticulously calculated object matching score to each proposal for ascertaining whether it aligns with a given novel object. In contrast to methods that solely focus on object semantics \cite{CNOS, ZeroPose}, we design the object matching scores considering three terms, including semantics, appearance and geometry. For each proposal, the first term assesses its semantic matching degree to the rendered templates of the object, while the second one further evaluates its appearance similarities to the best-matched template. The final term considers the matching degree based on geometry, such as object shape and size, by calculating the Intersection-over-Union (IoU) value between the bounding boxes of the proposal and the 2D projection of the object transformed by a rough pose estimate.

The Pose Estimation Model (PEM) is designed to calculate a 6D object pose for each identified proposal that matches the novel object. Initially, we formulate this pose estimation challenge as a partial-to-partial point matching problem between the sampled point sets of the proposal and the target object, considering the factors such as occlusions, segmentation inaccuracies, and sensor noises. To solve this problem, we propose a simple yet effective solution that involves the use of background tokens; specifically, for the two point sets, we learn to align their non-overlapped points with the background tokens in the feature space, and thus effectively establish an assignment matrix to build the necessary correspondence for predicting the object pose. Based on the design of background tokens, we further develop PEM with two point matching stages, \ie, \textbf{Coarse Point Matching} and \textbf{Fine Point Matching}. The first stage realizes sparse correspondence to derive an initial object pose, which is subsequently used to transform the point set of the proposal, enabling the learning of positional encodings. The second stage incorporates the positional encodings of the two point sets to inject the initial correspondence, and builds dense correspondence for estimating a more precise object pose. To effectively model dense interactions in the second stage, we propose an innovative design of Sparse-to-Dense Point Transformers, which realize interactions on the sparse versions of the dense features, and subsequently, distribute the enhanced sparse features back to the dense ones using Linear Transformers \cite{katharopoulos2020transformers, han2023flatten}.

For the two models of SAM-6D, ISM, built on SAM, does not require any network re-training or fine-tuning, while PEM is trained on the large-scale synthetic images of ShapeNet-Objects \cite{chang2015shapenet} and Google-Scanned-Objects \cite{downs2022google} datasets provided by \cite{MegaPose}. We evaluate SAM-6D on the seven core datasets of the BOP benchmark \cite{BOP}, including LM-O, T-LESS, TUD-L, IC-BIN, ITODD, HB, and YCB-V. The qualitative results are visualized in Fig. \ref{fig:vis}.  SAM-6D outperforms the existing methods on both tasks of instance segmentation and pose estimation of novel objects, thereby showcasing its robust generalization capabilities.

Our main contributions could be summarized as follows:

\begin{itemize}
  \item We propose a novel framework of SAM-6D, which realizes joint instance segmentation and pose estimation of novel objects from RGB-D images, and outperforms the existing methods on seven datasets of BOP benchmark.
  \item We leverage the zero-shot capacities of Segmentation Anything Model (SAM) to generate all possible proposals, and devise a novel object matching score to identify the proposals corresponding to novel objects.
  \item We approach pose estimation as a partial-to-partial point matching problem with a simple yet effective design of background tokens, and propose a two-stage point matching model for novel objects. The first stage realizes coarse point matching to derive initial object poses, which are then refined in the second stage of fine point matching using newly proposed Sparse-to-Dense Point Transformers.
\end{itemize}

\section{Related Work}
\label{sec:relatedwork}

\subsection{Segment Anything}

Segment Anything (SA) \cite{SAM}, is a promptable segmentation task that focuses on predicting valid masks for various types of prompts, \eg, points, boxes, text, and masks. To tackle this task, the authors propose a powerful segmentation model called Segment Anything Model (SAM), which comprises three components, including an image encoder, a prompt encoder and a mask decoder. SAM has demonstrated remarkable zero-shot transfer segmentation performance in real-world scenarios, including challenging situations such as medical images \cite{mazurowski2023segment, ma2023segment, zhang2023improving}, camouflaged objects \cite{tang2023can, ji2023sam}, and transparent objects \cite{han2023segment, ji2023segment}. Moreover, SAM has exhibited high versatility across numerous vision applications \cite{zhang2023survey}, such as image inpainting \cite{wang2023instructedit, xie2023edit, yu2023inpaint, liu2023internchat}, object tracking \cite{he2023scalable, yang2023track, zhang2023uvosam}, 3D detection and segmentation \cite{zhang2023sam3d, yang2023sam3d, cen2023sad}, and 3D reconstruction \cite{shen2023anything, wang2023inpaintnerf360, cen2023segment}.

Recent studies have also investigated semantically segmenting anything due to the critical role of semantics in vision tasks. 
Semantic Segment Anything (SSA) \cite{chen2023semantic} is proposed on top of SAM, aiming to assign semantic categories to the masks generated by SAM.
Both PerSAM \cite{PerSAM} and Matcher \cite{liu2023matcher} employ SAM to segment the object belonging to a specific category in a query image by searching for point prompts  with the aid of a reference image containing an object of the same category.
CNOS \cite{CNOS} is proposed to segment all instances of a given object model, which firstly generates mask proposals via SAM and subsequently filters out proposals with low feature similarities against object templates rendered from the object model.

For efficiency, FastSAM \cite{FastSAM} is proposed by utilizing instance segmentation networks with regular convolutional networks instead of visual transformers used in SAM. Additionally, MobileSAM \cite{MobileSAM} replaces the heavy encoder of SAM with a lightweight one through decoupled distillation.

\subsection{Pose Estimation of Novel Objects}

\vspace{0.05cm}
\noindent \textbf{Methods Based on Image Matching} Methods within this group \cite{okorn2021zephyr, liu2022gen6d, cai2022ove6d, shugurov2022osop, nguyen2022templates, pan2023learning, nguyen2023nope, MegaPose, nguyen2024gigaPose} often involve comparing object proposals to templates of the given novel objects, which are rendered with a series of object poses, to retrieve the best-matched object poses. For example, Gen6D \cite{liu2022gen6d}, OVE6D \cite{cai2022ove6d}, and GigaPose \cite{nguyen2024gigaPose} are designed to select the viewpoint rotations via image matching and then estimate the in-plane rotations to obtain the final estimates. MegaPose \cite{MegaPose} employs a coarse estimator to treat image matching as a classification problem, of which the recognized object poses are further updated by a refiner.

\vspace{0.05cm}
\noindent \textbf{Methods Based on Feature Matching} Methods within this group \cite{he2022fs6d, goodwin2022zero, sun2022onepose, he2022onepose++, fan2023pope, ZeroPose} align the 2D pixels or 3D points of the proposals with the object surface in the feature space \cite{sun2021loftr, huang2021predator}, thereby building correspondence to compute object poses. OnePose \cite{sun2022onepose} matches the pixel descriptors of proposals with the aggregated point descriptors of the point sets constructed by Structure from Motion (SfM) for 2D-3D correspondence, while OnePose++ \cite{he2022onepose++} further improves it with a keypoint-free SfM and a sparse-to-dense 2D-3D matching model. ZeroPose \cite{ZeroPose} realizes 3D-3D matching via geometric structures, and GigaPose \cite{nguyen2024gigaPose} establishes 2D-2D correspondence to regress in-plane rotation and 2D scale. Moreover, \cite{goodwin2022zero} introduces a zero-shot category-level 6D pose estimation task, along with a self-supervised semantic correspondence learning method. Unlike the above one-stage point matching work, the unique contributions in our Pose Estimation Model are: (a) a two-stage pipeline that boosts performance by incorporating coarse correspondence for finer matching, (b) an efficient design of background tokens to eliminate the need of optimal transport with iterative optimization \cite{qin2022geometric}, and (c) a Sparse-to-Dense Point Transformer to effectively model dense relationship.
\newpage

\section{Methodology of SAM-6D}
\label{sec:method}

We present \textbf{SAM-6D} for zero-shot 6D object pose estimation, which aims to detect all instances of a specific novel object, unseen during training, along with their 6D object poses in the RGB-D images. To realize the challenging task, SAM-6D breaks it down into two steps via two dedicated sub-networks, \ie, an \textbf{Instance Segmentation Model (ISM)} and a \textbf{Pose Estimation Model (PEM)}, to first segment all instances and then individually predict their 6D poses, as shown in Fig. \ref{fig:head}. We detail the architectures of ISM and PEM in Sec. \ref{subsec:ism} and Sec. \ref{subsec:pem}, respectively.


\subsection{Instance Segmentation Model}
\label{subsec:ism}

SAM-6D uses an Instance Segmentation Model (ISM) to segment the instances of a novel object $\mathcal{O}$. Given a cluttered scene, represented by an RGB image $\mathcal{I}$, ISM leverages the zero-shot transfer capabilities of Segment Anything Model (SAM) \cite{SAM} to generate all possible proposals $\mathcal{M}$. For each proposal $m\in\mathcal{M}$, ISM calculates an object matching score $s_m$ to assess the matching degree between $m$ and $\mathcal{O}$ in terms of semantics, appearance, and geometry. The matched instances with $\mathcal{O}$ can then be identified by simply setting a matching threshold $\delta_m$.

In this subsection, we initially provide a brief review of SAM in Sec. \ref{subsec:sam} and then explain the computation of the object matching score $s_m$ in Sec. \ref{subsec:matching-score}.

\vspace{-0.2cm}
\subsubsection{Preliminaries of Segment Anything Model}
\label{subsec:sam}
Given an RGB image $\mathcal{I}$, Segment Anything Model (SAM) \cite{SAM} realizes promptable segmentation with various types of prompts $\mathcal{P}_r$, \eg, points, boxes, texts, or masks. Specifically, SAM consists of three modules, including an image encoder $\Phi_{\text{Image}}$, a prompt encoder  $ \Phi_{\text{Prompt}}$, and a mask decoder $\Psi_{\text{Mask}}$, which could be formulated as follows:
\begin{equation}
    \mathcal{M}, \mathcal{C} = \Psi_{\text{Mask}} (\Phi_{\text{Image}}(I), \Phi_{\text{Prompt}}(\mathcal{P}_r)),
\label{eqn:rgb sam}
\end{equation}
where $\mathcal{M}$ and $\mathcal{C}$ denote the predicted proposals and the corresponding confidence scores, respectively. 

To realize zero-shot transfer, one can prompt SAM with evenly sampled 2D grids to yield all possible proposals, which can then be filtered based on confidence scores, retaining only those with higher scores, and applied to Non-Maximum Suppression to eliminate redundant detections.

\begin{figure*}[h]
  \centering
   \includegraphics[width=0.9\linewidth]{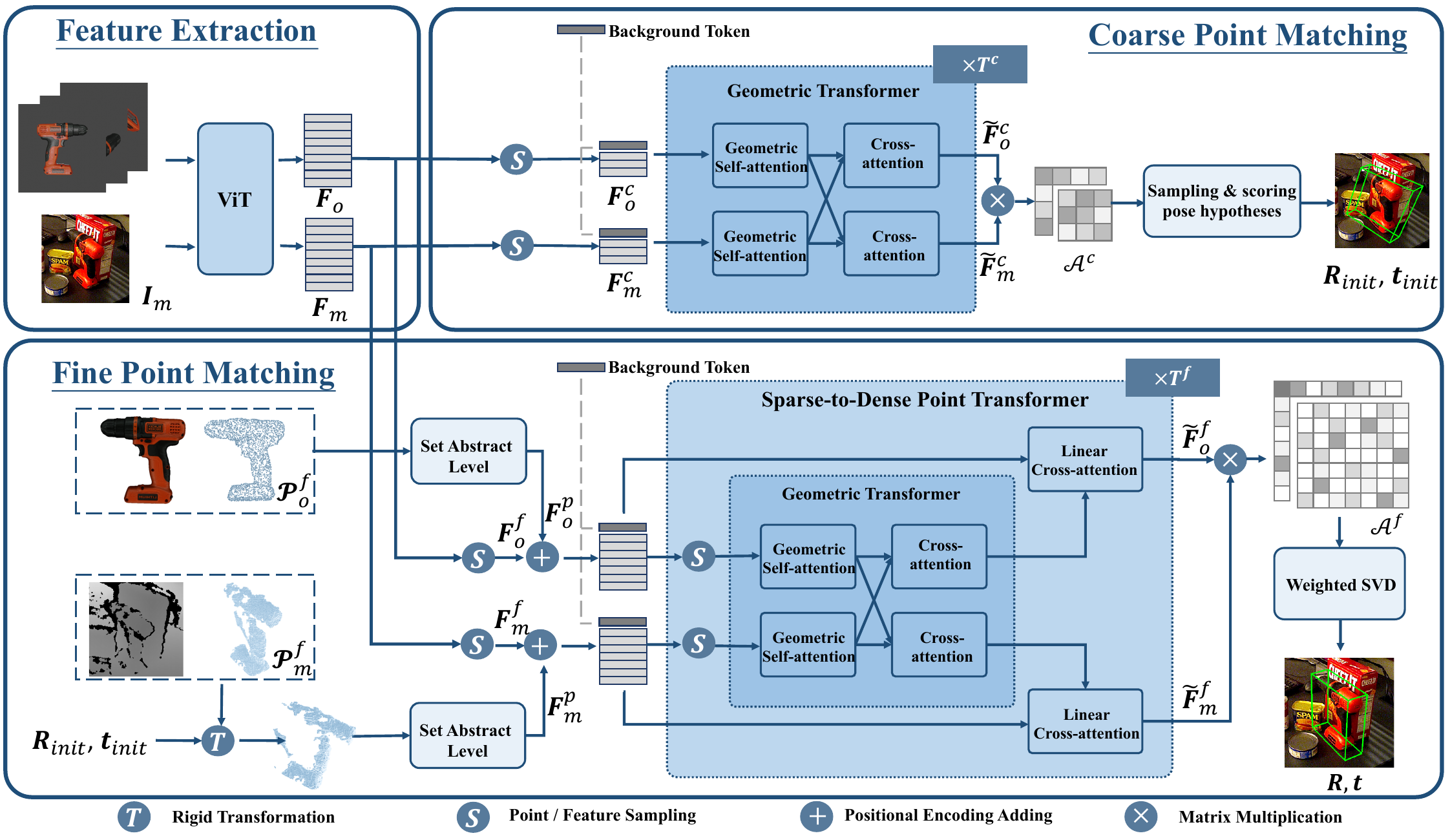}
   \vspace{-0.25cm}
   \caption{An illustration of Pose Estimation Model (PEM) of SAM-6D.}
   \vspace{-0.4cm}
   \label{fig:pem}
\end{figure*}

\vspace{-0.2cm}
\subsubsection{Object Matching Score}
\label{subsec:matching-score}

Given the proposals $\mathcal{M}$, the next step is to identify the ones that are matched with a specified object $\mathcal{O}$ by assigning each proposal $m \in \mathcal{M}$ with an object matching score $s_m$, which comprises three terms, each evaluating the matches in terms of semantics, appearance, and geometry, respectively. 

Following \cite{CNOS}, we sample $N_\mathcal{T}$ object poses in SE(3) space to render the templates $\{\mathcal{T}_k\}_{k=1}^{N_{\mathcal{T}}}$ of $\mathcal{O}$, which are fed into a pre-trained visual transformer (ViT) backbone \cite{ViT} of DINOv2 \cite{oquab2023dinov2}, resulting in the class embedding $\bm{f}^{cls}_{\mathcal{T}_{k}}$ and $N^{patch}_{\mathcal{T}_{k}}$ patch embeddings $\{\bm{f}^{patch}_{\mathcal{T}_{k}, i}\}_{i=1}^{N^{patch}_{\mathcal{T}_{k}}}$ of each template $\mathcal{T}_k$. For each proposal $m$, we crop the detected region out from $\mathcal{I}$, and resize it to a fixed resolution. The image crop is denoted as $\mathcal{I}_{m}$ and also processed through the same ViT to obtain the class embedding $\bm{f}^{cls}_{\mathcal{I}_{m}}$ and the patch embeddings $\{\bm{f}^{patch}_{\mathcal{I}_{m}, j}\}_{j=1}^{N^{patch}_{\mathcal{I}_{m}}}$, with $N^{patch}_{\mathcal{I}_{m}}$ denoting the number of patches within the object mask. Subsequently, we calculate the values of the individual score terms.

\vspace{0.1cm}
\noindent \textbf{Semantic Matching Score} 
We compute a semantic score $s_{sem}$ through the class embeddings by averaging the top K values from $\{\frac{<\bm{f}^{cls}_{\mathcal{I}_{m}}, \bm{f}^{cls}_{\mathcal{T}_{k}}>}{|\bm{f}^{cls}_{\mathcal{I}_{m}}|\cdot|\bm{f}^{cls}_{\mathcal{T}_{k}}|}\}_{k=1}^{N_{\mathcal{T}}}$ to establish a robust measure of semantic matching, with $<,>$ denoting an inner product. The template that yields the highest semantic value can be seen as the best-matched template, denoted as $\mathcal{T}_{best}$, and is used in the computation of the subsequent two scores.

\vspace{0.1cm}
\noindent \textbf{Appearance Matching Score}
Given $\mathcal{T}_{best}$, we compare $\mathcal{I}_{m}$ and $\mathcal{T}_{best}$ in terms of appearance using an appearance score $s_{appe}$, based on the patch embeddings, as follows:
\vspace{-0.2cm}
\begin{equation}
    s_{appe} = \frac{1}{N^{patch}_{\mathcal{I}_{m}}}\sum_{j=1}^{N^{patch}_{\mathcal{I}_{m}}} \max_{i=1,\dots,N^{patch}_{\mathcal{T}_{best}}} \frac{<\bm{f}^{patch}_{\mathcal{I}_{m},j}, \bm{f}^{patch}_{\mathcal{T}_{best},i}>}{|\bm{f}^{patch}_{\mathcal{I}_{m},j}|\cdot|\bm{f}^{patch}_{\mathcal{T}_{best},i}|}.
\label{eqn:s_ape}
\end{equation}
$s_{appe}$ is utilized to distinguish objects that are semantically similar but differ in appearance.

\vspace{0.1cm}
\noindent \textbf{Geometric Matching Score} In terms of geometry, we score the proposal $m$ by considering factors like object shapes and sizes. Utilizing the object rotation from $\mathcal{T}_{best}$ and the mean location of the cropped points of $m$, we have a coarse pose to transform the object $\mathcal{O}$, which is then projected onto the image to obtain a compact bounding box $\mathcal{B}_o$. Afterwards, the Intersection-over-Union (IoU) value between $\mathcal{B}_o$ and the bounding box $\mathcal{B}_m$ of $m$ is used as the geometric score $s_{geo}$:
\vspace{-0.2cm}
\begin{equation}
    s_{geo} = \frac{\mathcal{B}_m \bigcap \mathcal{B}_o}{\mathcal{B}_m \bigcup \mathcal{B}_o}.
\label{eqn:s_geo}
\end{equation}
The reliability of $s_{geo}$ is easily impacted by occlusions. We thus compute a visible ratio $r_{vis}$ to evaluate the confidence of $s_{geo}$, which is detailed in the supplementary materials.

By combining the above three score terms, the object matching score $s_m$ could be formulated as follows:
\vspace{-0.2cm}
\begin{equation}
    s_m = \frac{s_{sem} + s_{appe} + r_{vis} \cdot s_{geo}}{1+1+r_{vis}}. 
\end{equation}


\subsection{Pose Estimation Model}
\label{subsec:pem}

SAM-6D uses a Pose Estimation Model (PEM) to predict the 6D  poses of the proposals matched with the object $\mathcal{O}$.

For each object proposal $m$, PEM uses a strategy of point registration to predict the 6D pose w.r.t. $\mathcal{O}$. Denoting the sampled point set of $m$ as $\mathcal{P}_m \in \mathbb{R}^{N_m\times 3}$ with $N_m$ points and that of $\mathcal{O}$ as $\mathcal{P}_o \in \mathbb{R}^{N_o\times 3}$ with $N_o$ points, the goal is to solve an assignment matrix to present the partial-to-partial correspondence between $\mathcal{P}_m$ and $\mathcal{P}_o$. Partial-to-partial correspondence arises as $\mathcal{P}_o$ only partially matches $\mathcal{P}_m$ due to occlusions, and $\mathcal{P}_m$ may partially align with $\mathcal{P}_o$ due to segmentation inaccuracies and sensor noises. We propose to equip their respective point features $\bm{F}_m \in \mathbb{R}^{N_m\times C}$ and $\bm{F}_o\in \mathbb{R}^{N_o\times C}$ with learnable \textbf{Background Tokens}, denoted as $\bm{f}^{bg}_{m} \in \mathbb{R}^C$ and $\bm{f}^{bg}_{o}\in \mathbb{R}^C$, where $C$ is the number of feature channels. This simple design resolves the assignment problem of non-overlapped points in two point sets, and the partial-to-partial correspondence thus could be effectively built based on feature similarities. Specifically, we can first compute the attention matrix $\mathcal{A}$ as follows:
\begin{equation}
    \mathcal{A} = [\bm{f}^{bg}_{m}, \bm{F}_m] \times [\bm{f}^{bg}_{o}, \bm{F}_o]^T \in \mathbb{R}^{(N_m+1)\times (N_o+1)},
\label{eqn:S}
\end{equation}
and then obtain the soft assignment matrix $\tilde{\mathcal{A}}$ 
\begin{equation}
    \tilde{\mathcal{A}} = \text{Softmax}_{\text{row}}(\mathcal{A} / \tau) \cdot \text{Softmax}_{\text{col}}(\mathcal{A} / \tau),
\label{eqn:softS}
\end{equation}
where $\text{Softmax}_{\text{row}}()$ and $\text{Softmax}_{\text{col}}()$ denote Softmax operations executed along the row and column of the matrix, respectively. $\tau$ is a constant temperature. The values in each row of $\tilde{\mathcal{A}}$, excluding the first row associated with the background, indicate the matching probabilities of the point $\bm{p}_m \in \mathcal{P}_m$ aligning with background and the points in $\mathcal{P}_o$. Specifically, for $\bm{p}_m$, its corresponding point $\bm{p}_o \in \mathcal{P}_o$ can be identified by locating the index of the maximum score $\tilde{a} \in \tilde{\mathcal{A}}$ along the row; if this index equals zero, the embedding of $\bm{p}_m$ aligns with the background token, indicating it has no valid correspondence in $\mathcal{P}_o$. Once $\tilde{\mathcal{A}}$ is obtained, we can gather all the matched pairs $\{(\bm{p}_m, \bm{p}_o)\}$, along with their scores $\{\tilde{a}\}$, to compute the pose using weighted SVD.

Building on the above strategy with background tokens, PEM is designed in two point matching stages. For the proposal $m$ and the target object $\mathcal{O}$, the first stage involves \textbf{Coarse Point Matching} between their sparse point sets $\mathcal{P}_m^c$ and $\mathcal{P}_o^c$, while the second stage involves \textbf{Fine Point Matching} between their dense sets $\mathcal{P}_m^f$ and $\mathcal{P}_o^f$; we use the upper scripts `c' and `f' to indicate respective variables of these two stages. The aim of the first stage is to derive a coarse pose $\bm{R}_{init}$ and $\bm{t}_{init}$ from sparse correspondence. Then in the second stage, we use the initial pose to transform $\mathcal{P}_m^f$ for learning the positional encodings, and employ stacked \textbf{Sparse-to-Dense Point Transformers} to learn dense correspondence for a final pose $\bm{R}$ and $\bm{t}$. Prior to two point matching modules, we incorporate a \textbf{Feature Extraction} module to learn individual point features of $m$ and $\mathcal{O}$.  Fig. \ref{fig:pem} gives a detailed illustration of PEM.

\vspace{-0.3cm}
\subsubsection{Feature Extraction}
\vspace{-0.1cm}

The Feature Extraction module is employed to extract point-wise features $\bm{F}_m$ and $\bm{F}_o$ for the point sets $\mathcal{P}_m$ and $\mathcal{P}_o$ of the proposal $m$ and the given object $\mathcal{O}$, respectively.

Rather than directly extracting features from the discretized points $\mathcal{P}_m$, we utilize the visual transformer (ViT) backbone \cite{ViT} on its masked image crop $\mathcal{I}_m$ to capture patch-wise embeddings, which are then reshaped and interpolated to match the size of $\mathcal{I}_m$. Each point in $\mathcal{P}_m$ is assigned  with the corresponding pixel embedding, yielding $\bm{F}_m$.

We represent the object $\mathcal{O}$ with templates rendered from different camera views. All visible object pixels (points) are aggregated across views and sampled to create the point set $\mathcal{P}_o$. Corresponding pixel embeddings, extracted using the ViT backbone, are then used to form $\bm{F}_o$.

\vspace{-0.3cm}
\subsubsection{Coarse Point Matching}
\vspace{-0.1cm}

The Coarse Point Matching module is used to initialize a coarse object pose $\bm{R}_{init}$ and $\bm{t}_{init}$ by estimating a soft assignment matrix $\tilde{\mathcal{S}}^c$ between sparse versions of $\mathcal{P}_m$ and $\mathcal{P}_o$.

As shown in Fig. \ref{fig:pem}, we first sample a sparse point set $\mathcal{P}_m^c \in \mathbb{R}^{N^c_m\times 3}$ with $N^c_m$ points from $\mathcal{P}_m$, and $\mathcal{P}_o^c \in \mathbb{R}^{N^c_o\times 3}$ with $N^c_o$ points from $\mathcal{P}_o$, along with their respective sampled features $\bm{F}_m^{c}$ and $\bm{F}_o^{c}$. Then we concatenate $\bm{F}_m^{c}$ and $\bm{F}_o^{c}$ with learnable background tokens, and process them through $T^c$ stacked Geometric Transformers \cite{qin2022geometric}, each of which consists of a geometric self-attention for intra-point-set feature learning and a cross-attention for inter-point-set correspondence modeling. The processed features, denoted as $\tilde{\bm{F}}_m^{c}$ and $\tilde{\bm{F}}_o^{c}$, are subsequently used to compute the soft assignment matrix $\tilde{\mathcal{A}}^c$ based on (\ref{eqn:S}) and (\ref{eqn:softS}).

With $\tilde{\mathcal{A}}^c$, we obtain the matching probabilities between the overlapped points of $\mathcal{P}_m^c$ and $\mathcal{P}_o^c$, which can serve as the distribution to sample multiple triplets of point pairs and compute pose hypotheses \cite{ke2017efficient, haugaard2022surfemb}. We assign each pose hypothesis $\bm{R}_{\text{hyp}}$ and $\bm{t}_{\text{hyp}}$ a pose matching score $s_{\text{hyp}}$ as:
\vspace{-0.1cm}
\begin{equation}
    s_{\text{hyp}} = N^c_m \enspace / \sum_{\bm{p}_m^c \in \mathcal{P}_m^c} \min_{\bm{p}_o^c \in \mathcal{P}_o^c} || \bm{R}_{\text{hyp}}^T(\bm{p}_o^c - \bm{t}_{\text{hyp}}) - \bm{p}_m^c ||_2.
\end{equation}
Among the pose hypotheses, the one with the highest pose matching score is chosen as the initial pose $\bm{R}_{init}$ and $\bm{t}_{init}$ inputted into the next Fine Point Matching module.

\vspace{-0.2cm}
\subsubsection{Fine Point Matching}
\vspace{-0.1cm}

The Fine Point Matching module is utilized to build dense correspondence and estimate a more precise pose $\bm{R}$ and $\bm{t}$.

To build finer correspondence, we sample a dense point set $\mathcal{P}_m^f \in \mathbb{R}^{N^f_m\times 3}$ with $N^f_m$ points from $\mathcal{P}_m$, and $\mathcal{P}_o^f \in \mathbb{R}^{N^f_o\times 3}$ with $N^f_o$ points from $\mathcal{P}_o$, along with their respective sampled features $\bm{F}_m^{f}$ and $\bm{F}_o^{f}$. We then inject the initial correspondence, learned by the coarse point matching, through the inclusion of positional encodings. Specifically, we transform $\mathcal{P}_m^f$ with the coarse pose $\bm{R}_{init}$ and $\bm{t}_{init}$ and apply it to a multi-scale Set Abstract Level \cite{qi2017pointnet++} to learn the positional encodings $\bm{F}_m^{p}$; similarly, positional encodings $\bm{F}_o^p$ are also learned for $\mathcal{P}_o^f$. We then add  $\bm{F}_m^{p}$ and $\bm{F}_o^p$ to $\bm{F}_m^{f}$ and $\bm{F}_o^{f}$, concatenate each with a background token, and process them to yield $\tilde{\bm{F}}_m^{f}$ and $\tilde{\bm{F}}_o^{f}$, resulting in the soft assignment matrix $\tilde{\mathcal{A}}^f$ based on (\ref{eqn:S}) and (\ref{eqn:softS}). 

However, the commonly used transformers \cite{vaswani2017attention, qin2022geometric} incur a significant computational cost when learning dense point features. The recent Linear Transformers \cite{katharopoulos2020transformers, han2023flatten}, while being more efficient, exhibit less effective modeling of point interactions, since they implement attentions along the feature dimension. To address this, we propose a novel design of \textbf{Sparse-to-Dense Point Transformer} (SDPT), as shown in Fig. \ref{fig:pem}. Specifically, given two dense point features $\bm{F}_m^{f}$ and $\bm{F}_o^{f}$, SDPT first samples two sparse features from them and applies a Geometric Transformer \cite{qin2022geometric} to enhance their interactions, resulting in two improved sparse features, denoted as $\bm{F}_m^{f \prime}$ and $\bm{F}_o^{f \prime}$. SDPT then employs a Linear Cross-attention \cite{han2023flatten} to spread the information from $\bm{F}_m^{f \prime}$ to $\bm{F}_m^{f}$, treating the former as the key and value of the transformer, and the latter as the query. The same operations are applied to $\bm{F}_o^{f \prime}$ and $\bm{F}_o^{f}$ to update $\bm{F}_o^{f}$.

In Fine Point Matching, we stack $T^f$ SDPTs to model the dense correspondence and learn the soft assignment matrix $\tilde{\mathcal{A}}^f$. We note that, in each SDPT, the background tokens are consistently maintained in both sparse and dense point features. After obtaining $\tilde{\mathcal{A}}^f$, we search within $\mathcal{P}_o^f$ for the corresponding points to all foreground points in $\mathcal{P}_m^f$ along with the probabilities, building dense correspondence, and compute the final object pose  $\bm{R}$ and $\bm{t}$ via weighted SVD.

\section{Experiments}
\label{sec:exp}

\begin{table*}
    \centering
    \resizebox{0.97\textwidth}{!}{
    \begin{tabular}{l|c|ccc|ccccccc|c}
      \toprule
      \multirow{2}{*}{Method} & Segmentation & \multicolumn{3}{c|}{Object Matching Score} & \multicolumn{7}{c|}{BOP Dataset} & \multirow{2}{*}{Mean}\\
      \cline{3-12}
      & Model & $s_{sem}$ & $s_{appe}$ & $s_{geo}$ &   LM-O & T-LESS & TUD-L & IC-BIN & ITODD & HB & YCB-V & \\
      \midrule
      ZeroPose \cite{ZeroPose}  & SAM \cite{SAM} & - & - & -  &  34.4 & 32.7 & 41.4 & 25.1 & 22.4 & 47.8 & 51.9 & 36.5 \\
      CNOS  \cite{CNOS} & FastSAM \cite{FastSAM}& - & - & -   &  39.7 & 37.4 & 48.0 & 27.0 & 25.4 & 51.1 & 59.9 & 41.2 \\
      CNOS \cite{CNOS}  & SAM \cite{SAM} & - & -  &  - & 39.6 & 39.7 & 39.1 & 28.4 & 28.2 & 48.0 & 59.5 & 40.4 \\
      \midrule 
      \multirow{8}{*}{SAM-6D (Ours)}  & \multirow{4}{*}{FastSAM \cite{FastSAM}} & \checkmark & $\times$  & $\times$  & 39.5 & 37.6 & 48.7 & 25.7 & 25.3 & 51.2 & 60.2 & 41.2 \\
      &  & \checkmark & \checkmark  & $\times$   & 40.6 & 39.3 & 50.1 & 27.7 & 29.0 & 52.2 & 60.6 & 42.8  \\
      &  & \checkmark & $\times$  & \checkmark   & 40.4 & 41.4 & 49.7 & 28.2 & 30.1 & 54.0 & 61.1 & 43.6 \\
      &  & \checkmark & \checkmark  & \checkmark   & 42.2 & 42.0 & 51.7 & 29.3 & 31.9 & 54.8 & $\bm{62.1}$  & 44.9  \\
      \cline{2-13}
      & \multirow{4}{*}{SAM \cite{SAM}} & \checkmark & $\times$  & $\times$  & 43.4 & 39.1 & 48.2 & 33.3 & 28.8 & 55.1 & 60.3 &44.0 \\
      &  & \checkmark & \checkmark  & $\times$   & 44.4 & 40.8 & 49.8 & 34.5 & 30.0 & 55.7 & 59.5 &45.0 \\
      &  & \checkmark & $\times$  & \checkmark  & 44.0 & 44.7 & 54.8 & 33.8 & 31.5 & 58.3 & 59.9 & 46.7 \\
      &  & \checkmark & \checkmark  & \checkmark   & $\bm{46.0}$ & $\bm{45.1}$ & $\bm{56.9}$ & $\bm{35.7}$ & $\bm{33.2}$ & $\bm{59.3}$ & 60.5 & $\bm{48.1}$\\
      \bottomrule
    \end{tabular}}
    \vspace{-0.2cm}
    \caption{Instance segmentation results of different methods on the seven core datasets of the BOP benchmark \cite{BOP}. We report the mean Average Precision (mAP) scores at different Intersection-over-Union (IoU) values ranging from 0.50 to 0.95 with a step size of 0.05. }
    \vspace{-0.25cm}
    \label{tab:mask_sota}
\end{table*}

\begin{table*}
    \centering
    \resizebox{0.97\textwidth}{!}{
    \begin{tabular}{l|c|c|ccccccc|c}
      \toprule
      \multirow{2}{*}{Method} & \multirow{2}{*}{Input Type} & \multirow{2}{*}{Detection / Segmentation} & \multicolumn{7}{c|}{BOP Dataset} & \multirow{2}{*}{Mean} \\
      \cline{4-10}
      & & & LM-O & T-LESS & TUD-L & IC-BIN & ITODD & HB & YCB-V & \\
      \midrule
      \multicolumn{11}{c}{With Supervised Detection / Segmentation} \\
      \midrule
      MegaPose \cite{MegaPose} & RGB   & \multirow{6}{*}{MaskRCNN \cite{MaskRCNN}}  & 18.7 & 19.7 & 20.5 & 15.3 & 8.00 & 18.6 & 13.9 & 16.2 \\
      MegaPose$^\dagger$ \cite{MegaPose} & RGB   &  & 53.7 & 62.2 & 58.4 & 43.6 & 30.1 & 72.9 & 60.4 & 54.5 \\
      MegaPose$^\dagger$ \cite{MegaPose} & RGB-D & & 58.3 & 54.3 & 71.2 & 37.1 & 40.4 & 75.7 & 63.3 & 57.2 \\
      ZeroPose \cite{ZeroPose} & RGB-D & & 26.1 & 24.3 & 61.1 & 24.7 & 26.4 & 38.2 & 29.5 & 32.6 \\
      ZeroPose$^\dagger$ \cite{ZeroPose} & RGB-D & & 56.2 & 53.3 & $\bm{87.2}$ & 41.8 & $\bm{43.6}$ & 68.2 & 58.4 & 58.4  \\
      SAM-6D (Ours)  & RGB-D & &  $\bm{66.5}$ &  $\bm{66.0}$& 80.9 & $\bm{61.9}$ & 31.9 & $\bm{81.8}$ & $\bm{79.6}$ & $\bm{66.9}$ \\
      \midrule
      \multicolumn{11}{c}{With Zero-Shot Detection / Segmentation} \\
      \midrule
      ZeroPose \cite{ZeroPose} & RGB-D & \multirow{3}{*}{ZeroPose \cite{ZeroPose}}  &  26.0 & 17.8 & 41.2 & 17.7 & 38.0 & 43.9 & 25.7 & 25.7 \\
      ZeroPose$^\dagger$ \cite{ZeroPose}  & RGB-D & &  49.1 & 34.0 & 74.5 & 39.0 & 42.9 & 61.0 & 57.7 & 51.2  \\
      SAM-6D (Ours) & RGB-D &         & $\bm{63.5}$ & $\bm{43.0}$ & $\bm{80.2}$ & $\bm{51.8}$ & $\bm{48.4}$ & $\bm{69.1}$ & $\bm{79.2}$ & $\bm{62.2}$ \\
      \midrule
      MegaPose$^*$ \cite{MegaPose} & RGB   & \multirow{7}{*}{CNOS (FastSAM) \cite{CNOS}} & 22.9 & 17.7 & 25.8 & 15.2 & 10.8 & 25.1 & 28.1 & 20.8  \\
      MegaPose$^\dagger$$^*$ \cite{MegaPose} & RGB   &  & 49.9 & 47.7 & 65.3 & 36.7 & 31.5 & 65.4 & 60.1 & 50.9 \\
      MegaPose$^\dagger$$^*$ \cite{MegaPose} & RGB-D &  & 62.6 & 48.7 & $\bm{85.1}$ & 46.7 & 46.8 & 73.0 & 76.4 & 62.8 \\
      ZeroPose$^\dagger$$^*$  \cite{ZeroPose} & RGB-D &  & 53.8 & 40.0 & 83.5 & 39.2 & 52.1 & 65.3 & 65.3 & 57.0  \\
      GigaPose \cite{nguyen2024gigaPose} & RGB  & & 29.9 & 27.3 & 30.2 & 23.1 & 18.8 & 34.8 & 29.0 & 27.6 \\
      GigaPose$^\dagger$ \cite{nguyen2024gigaPose} & RGB & & 59.9 & $\bm{57.0}$ & 63.5 & 46.7 & 39.7 & 72.2 & 66.3 & 57.9 \\
      SAM-6D (Ours) & RGB-D &  &  $\bm{65.1}$ &  47.9 & 82.5 & $\bm{49.7}$ & $\bm{56.2}$ & $\bm{73.8}$ & $\bm{81.5}$ & $\bm{65.3}$ \\
      \midrule
      SAM-6D (Ours) & RGB-D & SAM-6D (FastSAM) & $\bm{66.7}$ & $\bm{48.5}$ & $\bm{82.9}$ & $\bm{51.0}$  & $\bm{57.2}$ & $\bm{73.6}$ & $\bm{83.4}$ & $\bm{66.2}$ \\
      \midrule
      SAM-6D (Ours) & RGB-D & SAM-6D (SAM) & $\bm{69.9}$ & $\bm{51.5}$ & $\bm{90.4}$ & $\bm{58.8}$  & $\bm{60.2}$ & $\bm{77.6}$ & $\bm{84.5}$ & $\bm{70.4}$ \\
      \bottomrule
    \end{tabular}}
    \vspace{-0.2cm}
    \caption{Pose estimation results of different methods on the seven core datasets of the BOP benchmark \cite{BOP}. We report the mean Average Recall (AR) among VSD, MSSD and MSPD, as introduced in Sec. \ref{sec:exp}. The symbol `$\dagger$' denotes the use of pose refinement proposed in \cite{MegaPose}. The symbol `$*$' denotes the results published on BOP leaderboard. Our used masks of MaskRCNN \cite{MaskRCNN} are provided by CosyPose \cite{labbe2020cosypose}.}
    \vspace{-0.35cm}
    \label{tab:pose_sota}
  \end{table*}

In this section, we conduct experiments to evaluate our proposed SAM-6D, which consists of an Instance Segmentation Model (ISM) and a Pose Estimation Model (PEM).

\vspace{0.1cm}
\noindent \textbf{Datasets} We evaluate our proposed SAM-6D on the seven core datasets of the BOP benchmark \cite{BOP}, including LM-O, T-LESS, TUD-L, IC-BIN, ITODD, HB, and YCB-V. PEM is trained on the large-scale synthetic ShapeNet-Objects \cite{chang2015shapenet} and Google-Scanned-Objects \cite{downs2022google} datasets provided by \cite{MegaPose}, with a total of $2,000,000$ images across $\sim 50,000$ objects.

\vspace{0.1cm}
\noindent \textbf{Implementation Details} For ISM, we follow \cite{CNOS} to utilize the default ViT-H SAM \cite{SAM} or FastSAM \cite{FastSAM} for proposal generation, and the default ViT-L model of DINOv2 \cite{DINOv2} to extract class and patch embeddings. For PEM, we set $N_m^c=N_o^c=196$ and $N_m^f=N_o^f=2048$, and use InfoNCE loss \cite{oord2018representation} to supervise the learning of attention matrices (\ref{eqn:S}) for both matching stages. We use ADAM to train PEM with a total of 600,000 iterations; the learning rate is initialized as 0.0001, with a cosine annealing schedule used, and the batch size is set as 28. For each object, we use two rendered templates for training PEM. During evaluation, we follow \cite{CNOS} and use 42 templates for both ISM and PEM.

\vspace{0.1cm}
\noindent \textbf{Evaluation Metrics} For instance segmentation, we report the mean Average Precision (mAP) scores at different Intersection-over-Union (IoU) thresholds ranging from 0.50 to 0.95 with a step size of 0.05. For pose estimation, we report the mean Average Recall (AR) w.r.t three error functions, \ie, Visible Surface Discrepancy (VSD), Maximum Symmetry-Aware Surface Distance (MSSD) and Maximum Symmetry-Aware Projection Distance (MSPD). For further details about these evaluation metrics, please refer to \cite{BOP}.

\subsection{Instance Segmentation of Novel Objects}
\vspace{-0.1cm}

We compare our ISM of SAM-6D with ZeroPose \cite{ZeroPose} and CNOS \cite{CNOS}, both of which score the object proposals in terms of semantics solely, for instance segmentation of novel objects. The quantitative results are presented in Table \ref{tab:mask_sota}, demonstrating that our ISM, built on the publicly available foundation models of SAM \cite{SAM} / FastSAM \cite{FastSAM} and ViT (pre-trained by DINOv2 \cite{DINOv2}), delivers superior results without the need for network re-training or finetuning. Note that our baseline with only semantic matching score $s_{sem}$, whether based on SAM or FastSAM \cite{FastSAM}, aligns precisely with the method of CNOS; the only difference is that we adjust the hyperparameters of SAM to generate more proposals for scoring. Further enhancements to our baselines are achieved via the inclusion of appearance and geometry matching scores, \ie, $s_{appe}$ and $s_{geo}$, as verified in Table \ref{tab:mask_sota}. Qualitative results of ISM are visualized in Fig. \ref{fig:vis}.

\subsection{Pose Estimation of Novel Objects}

\subsubsection{Comparisons with Existing Methods}
\vspace{-0.1cm}

We compare our PEM of SAM-6D with the representative methods, including MegaPose \cite{MegaPose}, ZeroPose \cite{ZeroPose}, and GigaPose \cite{nguyen2024gigaPose}, for pose estimation of novel objects. Quantitative comparisons, as presented in Table \ref{tab:ba-transformer}, show that our PEM, without the time-intensive render-based refiner \cite{MegaPose}, outperforms the existing methods under various mask predictions. Importantly, the mask predictions from our ISM significantly enhance the performance of PEM, compared to other mask predictions, further validating the advantages of ISM. Qualitative results of PEM are visualized in Fig. \ref{fig:vis}.

\vspace{-0.3cm}
\subsubsection{Ablation Studies and Analyses} 
\vspace{-0.1cm}

We conduct ablation studies on the YCB-V dataset to evaluate the efficacy of individual designs in PEM, with the mask predictions generated by ISM based on SAM.

\vspace{0.05cm}
\noindent \textbf{Efficacy of Background Tokens} We address the partial-to-partial point matching issue through a simple yet effective design of background tokens. Another existing solution is the use of optimal transport \cite{qin2022geometric}  with iterative optimization, which, however, is time-consuming. The two solutions are compared in Table \ref{tab:ba-bgtoken}, which shows that our PEM with background tokens achieves results comparable to optimal transport, but with a faster inference speed. As the density of points for matching increases, optimal transport requires more time to derive the assignment matrices.

\vspace{0.05cm}
\noindent \textbf{Efficacy of Two Point Matching Stages} With the background tokens, we design PEM with two stages of point matching via a Coarse Point Matching module and a Fine Point Matching module. Firstly, we validate the effectiveness of the Fine Point Matching module, which effectively improves the results of the coarse module, as verified in Table \ref{tab:ba-coarse-point-matching}. Further, we evaluate the effectiveness of the Coarse Point Matching module by removing it from PEM. In this case, the point sets of object proposals are not transformed and are directly used to learn the positional encodings in the fine module. The results, presented in Table \ref{tab:ba-coarse-point-matching}, indicate that the removal of Coarse Point Matching significantly degrades the performance, which may be attributed to the large distance between the sampled point sets of the proposals and target objects, as no initial poses are provided.

\vspace{0.05cm}

\noindent \textbf{Efficacy of Sparse-to-Dense Point Transformers} We design Sparse-to-Dense Point Transformers (SDPT) in the Fine Point Matching module to manage dense point interactions. Within each SDPT, Geometric Transformers \cite{qin2022geometric} is employed to learn the relationships between sparse point sets, which are then spread to the dense ones via Linear Transformers \cite{katharopoulos2020transformers}. We conduct experiments on either Geometric Transformers using sparse point sets with 196 points or Linear Transformers using dense point sets with 2048 points. The results, presented in Table \ref{tab:ba-transformer}, indicate inferior performance compared to using our SDPTs. This is because Geometric Transformers struggle to handle dense point sets due to high computational costs, whereas Linear Transformers prove to be ineffective in modeling dense correspondence with attention along the feature dimension.

\begin{table}
  \centering
  \resizebox{0.37\textwidth}{!}{
  \begin{tabular}{l|c|c}
    \toprule
      & AR & Time (s) \\
    \midrule
     PEM with Optimal Transport & 81.4 & 4.31 \\
     PEM with Background Tokens & $\bm{84.5}$ & $\bm{1.36}$\\
    \bottomrule
  \end{tabular} }
  \vspace{-0.2cm}
  \caption{Quantitative results of Optimal Transport \cite{qin2022geometric} and our design of Background Tokens in the Pose Estimation Model on YCB-V. The reported time is the average per-image processing time of pose estimation across the entire dataset on a server with a GeForce RTX 3090 GPU.}
  \vspace{-0.15cm}
  \label{tab:ba-bgtoken}
\end{table}

\begin{table}
  \centering
  \resizebox{0.38\textwidth}{!}{
  \begin{tabular}{c|c|c}
    \toprule
    Coarse Point Matching & Fine Point Matching  &  AR \\
    \midrule
    \checkmark & $\times$   & 77.6 \\
    $\times$  & \checkmark & 40.2\\
    \checkmark  & \checkmark & $\bm{84.5}$\\
    \bottomrule
  \end{tabular} }
  \vspace{-0.2cm}
  \caption{Ablation studies on the the strategy of two point matching stages in the Pose Estimation Model on YCB-V.}
  \vspace{-0.15cm}
  \label{tab:ba-coarse-point-matching}
\end{table}

\begin{table}
    \centering
    \resizebox{0.42\textwidth}{!}{
    \begin{tabular}{l|c|c}
      \toprule
      Transformer & $\#$Point & AR \\
      \midrule
      Geometric Transformer \cite{qin2022geometric} & 196 & 81.7\\
      Linear Transformer \cite{katharopoulos2020transformers} & 2048 & 78.4 \\
      Sparse-to-Dense Point Transformer & 196 $\rightarrow$ 2048 & $\bm{84.5}$\\
      \bottomrule
    \end{tabular}}
    \vspace{-0.2cm}
    \caption{Quantitative comparisons among various types of transformers employed in the Fine Point Matching module of the Pose Estimation Model on YCB-V.}
    \vspace{-0.15cm}
    \label{tab:ba-transformer}
\end{table}

\begin{table}
  \centering
  \resizebox{0.47\textwidth}{!}{
  \begin{tabular}{c|cc|c}
    \toprule
    \multirow{2}{*}{Segmentation Model} & \multicolumn{3}{c}{Time (s)} \\
    \cline{2-4}
     & Instance Segmentation & Pose Estimaiton  &  All \\
    \midrule
    FastSAM \cite{FastSAM} & 0.45 &  0.98  & 1.43 \\
    SAM \cite{SAM}  & 2.80 & 1.57 & 4.37 \\
    \bottomrule
  \end{tabular} }
  \vspace{-0.2cm}
  \caption{Runtime of SAM-6D with different segmentation models. The reported time is the average per-image processing time across the seven core datasets of BOP benchmark on a server with a GeForce RTX 3090 GPU.}
  \vspace{-0.4cm}
  \label{tab:running-time}
\end{table}

\vspace{-0.1cm}
\subsection{Runtime Analysis}
\vspace{-0.1cm}

We conduct evaluation on a server with a GeForce RTX 3090 GPU, and report in Table \ref{tab:running-time} the runtime  averaged on the seven core datasets of BOP benchmark, indicating the efficiency of SAM-6D which avoids the use of time-intensive render-based refiners. We note that SAM-based method takes more time on pose estimation than FastSAM-based one, due to more object proposals generated by SAM.

\vspace{-0.1cm}
\section{Conclusion}
\vspace{-0.1cm}

In this paper, we take Segment Anything Model (SAM) as an advanced starting point for zero-shot 6D object pose estimation, and present a novel framework, named SAM-6D, which comprises an Instance Segmentation Model (ISM) and a Pose Estimation Model (PEM) to accomplish the task in two steps. ISM utilizes SAM to segment all potential object proposals and assigns each of them an object matching score in terms of semantics, appearance, and geometry. PEM then predicts the object pose for each proposal by solving a partial-to-partial point matching problem through two stages of Coarse Point Matching and Fine Point Matching. The effectiveness of SAM-6D is validated on the seven core datasets of BOP benchmark, where SAM-6D significantly outperforms existing methods.

{
    \small
    \bibliographystyle{ieeenat_fullname}
    \bibliography{main}
}

\clearpage
\setcounter{section}{0}
\setcounter{page}{1}
\maketitlesupplementary
\renewcommand\thesection{\Alph{section}}

\noindent \textbf{CONTENT:}

\begin{itemize}
    \item \S \textbf{\ref{sec:supp_ism}. Supplementary Material for Instance Segmentation Model}
    \begin{itemize}
        \item[\textbullet] \S \textbf{\ref{subsec:supp_ism_vis_ratio}. Visible Ratio for Geometric Matching Score}
        \item[\textbullet] \S \textbf{\ref{subsec:supp_ism_tem_select}. Template Selection for Object Matching}
        \item[\textbullet] \S \textbf{\ref{subsec:supp_ism_super-param}. Hyperparameter Settings}
        \item[\textbullet] \S \textbf{\ref{subsec:supp_ism_quan}. More Quantitative Results}
        \begin{itemize}
            \item[\textbullet] \S \ref{subsec:supp_ism_quan_det}. Detection Results
            \item[\textbullet] \S \ref{subsec:supp_ism_quan_model_size}. Effects of Model Sizes
        \end{itemize}
        \item[\textbullet] \S \textbf{\ref{subsec:supp_ism_qual}. More Qualitative Results}
        \begin{itemize}
            \item[\textbullet] \S \ref{subsec:supp_ism_qual_appe}. Qualitative Comparisons on Appearance Matching Score
            \item[\textbullet] \S \ref{subsec:supp_ism_qual_geo}. Qualitative Comparisons on Geometric Matching Score
            \item[\textbullet] \S \ref{subsec:supp_ism_qual_vis}. More Qualitative Comparisons with Existing Methods
        \end{itemize}
    \end{itemize}
    
    \vspace{0.2cm}
    \item \S \textbf{\ref{sec:supp_pem}. Supplementary Material for Pose Estimation Model}
    \begin{itemize}
        \item[\textbullet] \S \textbf{\ref{subsec:supp_pem_arch}. Network Architectures and Specifics}
        \begin{itemize}
            \item[\textbullet] \S \ref{subsec:supp_pem_arch_1}. Feature Extraction
            \item[\textbullet] \S \ref{subsec:supp_pem_arch_2}. Coarse Point Matching
            \item[\textbullet] \S \ref{subsec:supp_pem_arch_3}. Fine Point Matching
        \end{itemize}
        \item[\textbullet] \S \textbf{\ref{subsec:supp_pem_obj}. Training Objectives}
        \item[\textbullet] \S \textbf{\ref{subsec:supp_pem_quan}. More Quantitative Results}
        \begin{itemize}
            \item[\textbullet] \S \ref{subsec:supp_pem_quan_view}. Effects of The View Number of Templates
            \item[\textbullet] \S \ref{subsec:supp_pem_quan_ove6d}. Comparisons with OVE6D
        \end{itemize}
        \item[\textbullet] \S \textbf{\ref{subsec:supp_pem_qual}. More Qualitative Comparisons with Existing Methods}
    \end{itemize}

\end{itemize}

\section{Supplementary Material for\\Instance Segmentation Model}
\label{sec:supp_ism}

\subsection{Visible Ratio for Geometric Matching Score}
\label{subsec:supp_ism_vis_ratio}

In the Instance Segmentation Model (ISM) of our SAM-6D,  we introduce a visible ratio $r_{vis}$ to weight the reliability of the geometric matching score $s_{geo}$.  
Specifically, given an RGB crop $\mathcal{I}_{m}$ of a proposal $m$ and the best-matched template $\mathcal{T}_{best}$ of the target object $\mathcal{O}$, along with their patch embeddings  $\{\bm{f}^{patch}_{\mathcal{I}_{m}, j}\}_{j=1}^{N^{patch}_{\mathcal{I}_{m}}}$ and $\{\bm{f}^{patch}_{\mathcal{T}_{best}, i}\}_{i=1}^{N^{patch}_{\mathcal{T}_{best}}}$, $r_{vis}$ is calculated as the ratio of patches in $\mathcal{T}_{best}$ that can find a corresponding patch in $\mathcal{I}_{m}$, estimating the occlusion degree of $\mathcal{O}$ in $\mathcal{I}_{m}$. We can formulate the calculation of visible ratio $r_{vis}$ as follows:
\begin{equation}
    r_{vis} = \frac{1}{N^{patch}_{\mathcal{T}_{best}}} \sum_{i=1}^{N^{patch}_{\mathcal{T}_{best}}} r_{vis, i},
\end{equation}
where 
\begin{equation}
r_{vis, i}  = \left\{\begin{matrix}
  0 & \text{if}\enspace s_{vis,i} < \delta_{vis}  \\
  1 & \text{if}\enspace s_{vis,i} \ge \delta_{vis}
\end{matrix}\right.,\notag
\end{equation}
and
\begin{equation}
    s_{vis,i} = \max_{j=1,\dots,N^{patch}_{\mathcal{I}_{m}}} \frac{<\bm{f}^{patch}_{\mathcal{I}_{m},j}, \bm{f}^{patch}_{\mathcal{T}_{best},i}>}{|\bm{f}^{patch}_{\mathcal{I}_{m},j}|\cdot|\bm{f}^{patch}_{\mathcal{T}_{best},i}|}.
\end{equation}
The constant threshold $\delta_{vis}$ is empirically set as 0.5 to determine whether the patches in $\mathcal{T}_{best}$ are occluded.

\subsection{Template Selection for Object Matching}
\label{subsec:supp_ism_tem_select}
For each given target object, we follow \cite{CNOS} to first sample $42$ well-distributed viewpoints defined by the icosphere primitive of Blender. Corresponding to these viewpoints, we select 42 fully visible object templates from the Physically-based Rendering (PBR) training images of the BOP benchmark \cite{BOP} by cropping regions and masking backgrounds using the ground truth object bounding boxes and masks, respectively. These cropped and masked images then serve as the templates of the target object, which are used to calculate the object matching scores for all generated proposals. It's noted that these 42 templates can also be directly rendered using the pre-defined viewpoints.

\subsection{Hyperparameter Settings}
\label{subsec:supp_ism_super-param}
In the paper, we use SAM \cite{SAM} based on ViT-H or FastSAM based on YOLOv8x as the segmentation model, and ViT-L of DINOv2 \cite{DINOv2} as the description model. We utilize the publicly available codes for autonomous segmentation from SAM and FastSAM, with the hyperparameter settings displayed in Table \ref{tab:ism_param}.

\begin{table}
    \centering
    \begin{tabular}{l|c}
      \toprule
      Hyperparameter & Setting \\
      \midrule
      \multicolumn{2}{c}{(a) SAM \cite{SAM}} \\
      \midrule
      point\_per\_size & 32 \\
      pred\_iou\_thresh & 0.88 \\
      stability\_score\_thresh & 0.85 \\
      stability\_score\_offset & 1.0 \\
      box\_nms\_thresh & 0.7 \\
      crop\_n\_layer & 0 \\
      point\_grids & None \\
      min\_mask\_region\_area & 0 \\
      \midrule
      \multicolumn{2}{c}{(b) FastSAM \cite{FastSAM}} \\
      \midrule
      iou & 0.9 \\
      conf & 0.05 \\
      max\_det & 200 \\
      \bottomrule
    \end{tabular} 
    \caption{Hyperparameter Settings of (a) SAM \cite{SAM} and (b) FastSAM \cite{FastSAM} in their publicly available codes for autonomous segmentation.}
    \label{tab:ism_param}
\end{table}

\subsection{More Quantitative Results}
\label{subsec:supp_ism_quan}

\subsubsection{Detection Results}
\label{subsec:supp_ism_quan_det}

We compare our Instance Segmentation Model (ISM) with ZeroPose \cite{ZeroPose} and CNOS  \cite{CNOS} in terms of 2D object detection in Table \ref{tab:detect_sota}, where our ISM outperforms both methods owing to the meticulously crafted design of object matching score. 
\begin{table*}
    \centering
    \begin{tabular}{@{}l|c|ccccccc|c@{}}
      \toprule
      \multirow{2}{*}{Method} & Segmentation & \multicolumn{7}{c|}{BOP Dataset} & \multirow{2}{*}{Mean}\\
      \cline{3-9}
      & Model &   LM-O & T-LESS & TUD-L & IC-BIN & ITODD & HB & YCB-V & \\
      \midrule
      ZeroPose \cite{ZeroPose}  & SAM \cite{SAM}   & 36.7 & 30.0 & 43.1 & 22.8 & 25.0 & 39.8 & 41.6 & 34.1 \\
      CNOS  \cite{CNOS} & FastSAM \cite{FastSAM}  & 43.3 & 39.5 & 53.4 & 22.6 & 32.5 & 51.7 & 56.8 & 42.8 \\
      CNOS \cite{CNOS}  & SAM \cite{SAM} & 39.5 & 33.0 & 36.8 & 20.7 & 31.3 & 42.3 & 49.0 & 36.1 \\
      \midrule 
      SAM-6D & FastSAM \cite{FastSAM}  & 46.3 & $\bm{45.8}$ & $\bm{57.3}$ & 24.5 & $\bm{41.9}$ & $\bm{55.1}$ & $\bm{58.9}$ & $\bm{47.1}$  \\
      SAM-6D & SAM \cite{SAM}  & $\bm{46.6}$ & 43.7 & 53.7 & $\bm{26.1}$ & 39.3 & 53.1 & 51.9 & 44.9 \\
      \bottomrule
    \end{tabular}
    \caption{Object Detection results of different methods on the seven core datasets of the BOP benchmark \cite{BOP}. We report the mean Average Precision (mAP) scores at different Intersection-over-Union (IoU) values ranging from 0.50 to 0.95 with a step size of 0.05.}
    \label{tab:detect_sota}
\end{table*}

\subsubsection{Effects of Model Sizes} 
\label{subsec:supp_ism_quan_model_size}
We draw a comparison across different model sizes for both segmentation and description models on YCB-V dataset in Table \ref{tab:seg_model_size}, which indicates a positive correlation between larger model sizes and higher performance for both models.

\begin{table}
    \centering
    \resizebox{0.4\textwidth}{!}{
    \begin{tabular}{cc|cc|c}
      \toprule
       \multicolumn{2}{c|}{Segmentation Model} & \multicolumn{2}{c|}{Description Model} & \multirow{2}{*}{AP} \\
       \cline{1-4}
       Type & $\#$Param & Type & $\#$Param & \\
      \midrule
      \multirow{2}{*}{FastSAM-s} & \multirow{2}{*}{23 M} & ViT-S & 21 M & 43.1 \\
       & & ViT-L & 300 M & 54.0 \\
       \hline
      \multirow{2}{*}{FastSAM-x} & \multirow{2}{*}{138 M} & ViT-S & 21 M & 48.9 \\
       & & ViT-L & 300 M & 62.0 \\
      \midrule \midrule
      \multirow{2}{*}{SAM-B} & \multirow{2}{*}{357 M} & ViT-S & 21 M & 44.0 \\
       & & ViT-L & 300 M & 55.8 \\
       \hline
      \multirow{2}{*}{SAM-L} & \multirow{2}{*}{1,188 M} & ViT-S & 21 M & 47.2 \\
       & & ViT-L & 300 M & 59.8 \\
       \hline
      \multirow{2}{*}{SAM-H} & \multirow{2}{*}{2,437 M} & ViT-S & 21 M & 47.1 \\
       & & ViT-L & 300 M & 60.5 \\      
      \bottomrule
    \end{tabular}}
    \caption{Quantitative comparisons on the model sizes of both segmentation and description models on YCB-V. We report the mean Average Precision (mAP) scores at different Intersection-over-Union (IoU) values ranging from 0.50 to 0.95 with a step size of 0.05. }
    \label{tab:seg_model_size}
\end{table}

\begin{table}
    \centering
    \resizebox{0.45\textwidth}{!}{
    \begin{tabular}{l|c|c|c}
      \toprule
      \multirow{2}{*}{Method} & Segmentation & \multirow{2}{*}{Server} & \multirow{2}{*}{Time (s)} \\
      & Model & & \\
      \midrule
      CNOS \cite{CNOS} & \multirow{3}{*}{FastSAM \cite{FastSAM}} & Tesla V100 & 0.22\\
      CNOS \cite{CNOS} &  & DeForce RTX 3090 & 0.23 \\
      SAM-6D  &  & DeForce RTX 3090 & 0.45 \\
      \midrule
      CNOS \cite{CNOS} & \multirow{3}{*}{SAM \cite{SAM}} & Tesla V100 & 1.84 \\
      CNOS \cite{CNOS} &  &  DeForce RTX 3090 & 2.35 \\
      SAM-6D  &  &  DeForce RTX 3090 & 2.80 \\
      \bottomrule
    \end{tabular}}
    \caption{Runtime comparisons of different methods for instance segmentation of novel objects.  The reported time is the average per-image processing time across the seven core datasets of the BOP benchmark \cite{BOP}.}
    \label{tab:seg_runtime}
\end{table}

\subsection{More Qualitative Results}
\label{subsec:supp_ism_qual}

\subsubsection{Qualitative Comparisons on Appearance Matching Score} 
\label{subsec:supp_ism_qual_appe}
\begin{figure}[t]
  \centering
   \includegraphics[width=0.98\linewidth]{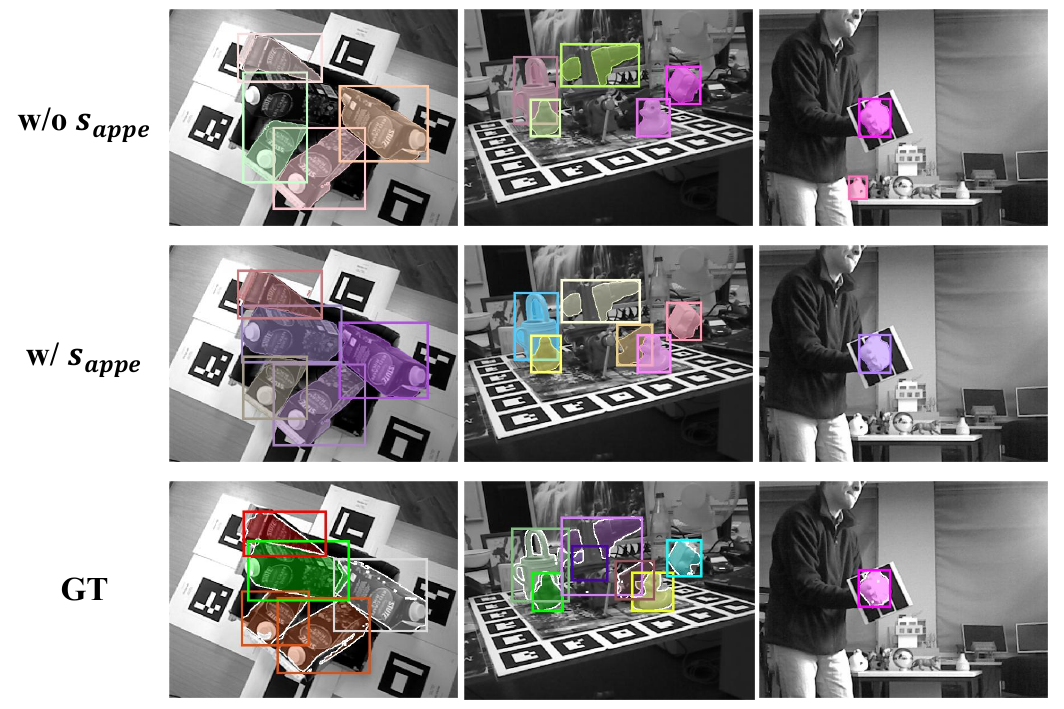}
   \caption{Qualitative results of our Instance Segmentation Model with or without the appearance matching score $s_{appe}$.}
   \label{fig:seg_appe}
\end{figure}

\vspace{-0.2cm}
We visualize the qualitative comparisons of the appearance matching score $s_{appe}$ in Fig. \ref{fig:seg_appe} to show its advantages in scoring the proposals w.r.t. a given object in terms of appearance.

\subsubsection{Qualitative Comparisons on Geometric Matching Score} 
\label{subsec:supp_ism_qual_geo}
\begin{figure}[t]
  \centering
   \includegraphics[width=0.98\linewidth]{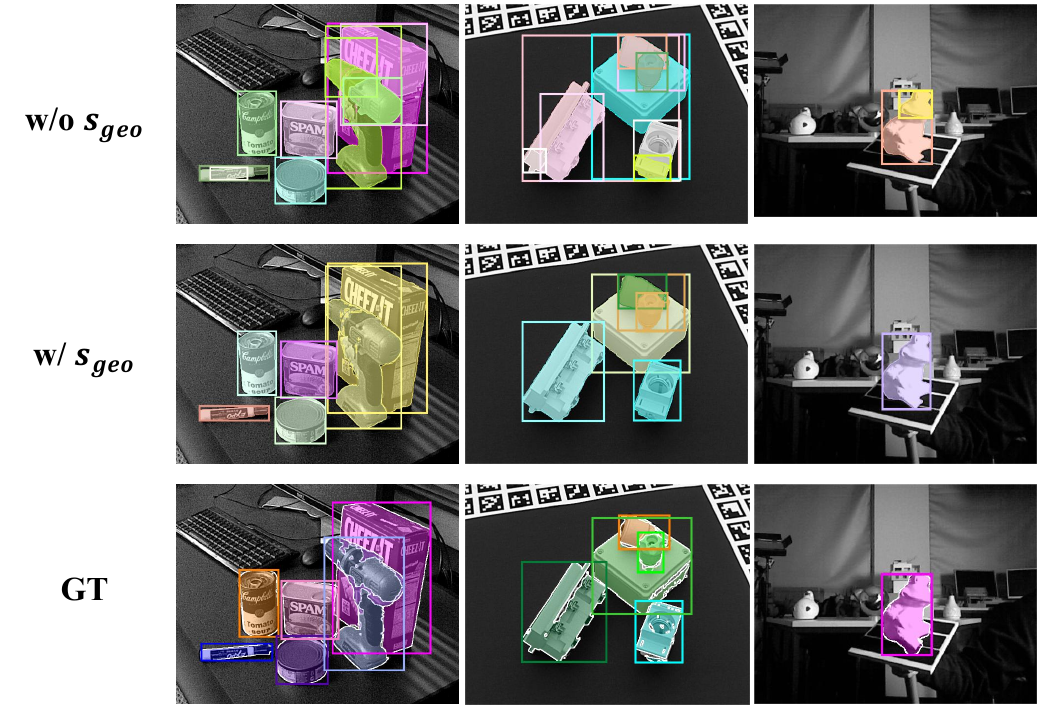}
   \caption{Qualitative results of our Instance Segmentation Model with or without the geometric matching score $s_{geo }$.}
   \label{fig:seg_geo}
\end{figure}

\vspace{-0.2cm}
We visualize the qualitative comparisons of the geometric matching score $s_{geo}$ in Fig. \ref{fig:seg_geo} to show its advantages in scoring the proposals w.r.t. a given object in terms of geometry, \eg, object shapes and sizes.

\subsubsection{More Qualitative Comparisons with Existing Methods}
\label{subsec:supp_ism_qual_vis}
\begin{figure*}[t]
  \centering
   \includegraphics[width=1.0\linewidth]{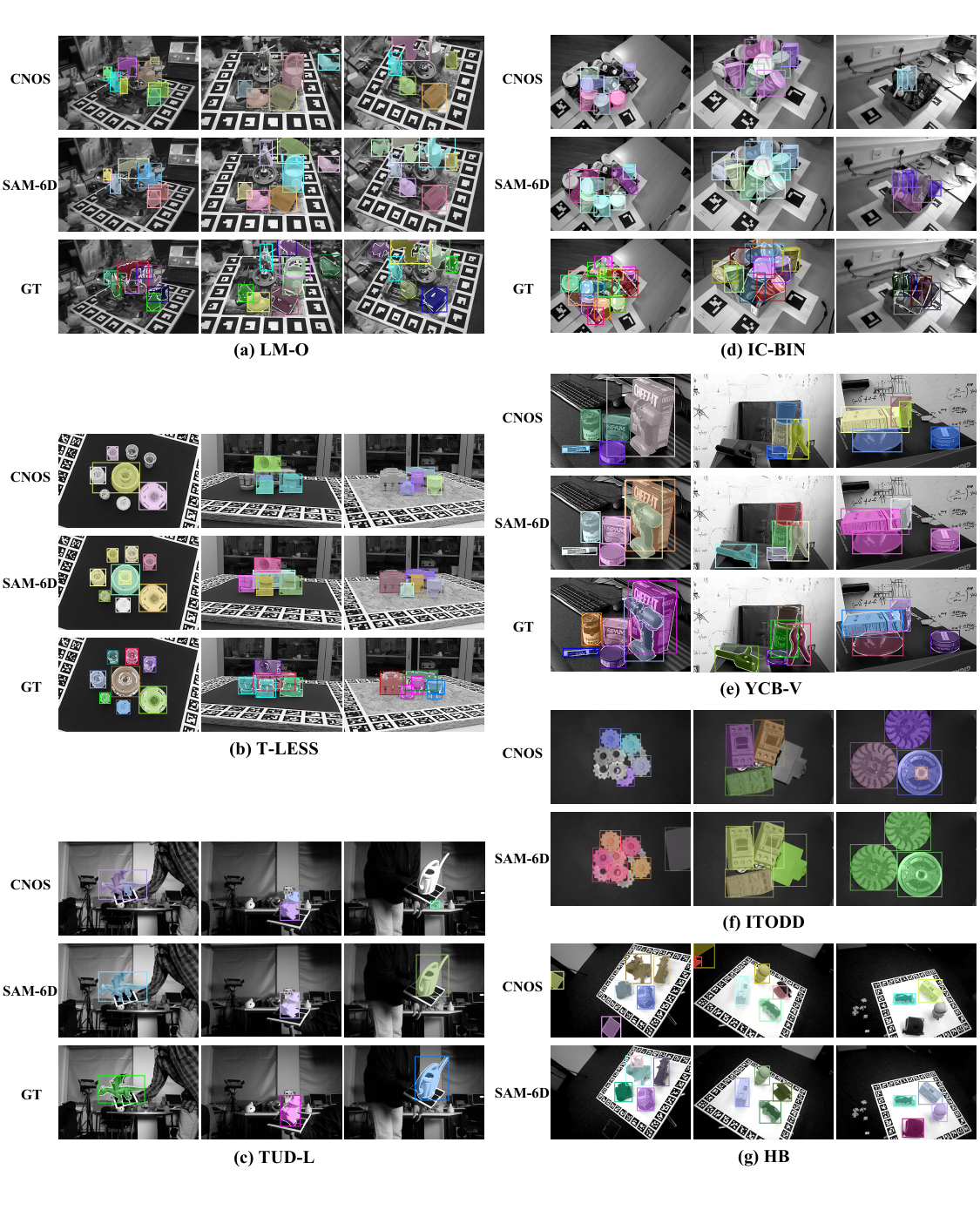}
   \caption{Qualitative results on the seven core datasets of the BOP benchmark \cite{BOP} for instance segmentation of novel objects.}
   \label{fig:seg_7dataset}
\end{figure*}

To illustrate the advantages of our Instance Segmentation Model (ISM), we visualize in Fig. \ref{fig:seg_7dataset} the qualitative comparisons with CNOS \cite{CNOS} on all the seven core datasets of the BOP benchmark \cite{BOP} for instance segmentation of novel objects. For reference, we also provide the ground truth masks, except for the ITODD and HB datasets, as their ground truths are not available.

\section{Supplementary Material for\\Pose Estimation Model}
\label{sec:supp_pem}

\subsection{Network Architectures and Specifics}
\label{subsec:supp_pem_arch}

\subsubsection{Feature Extraction}
\label{subsec:supp_pem_arch_1}

\begin{figure*}[t]
  \centering
   \includegraphics[width=1.0\linewidth]{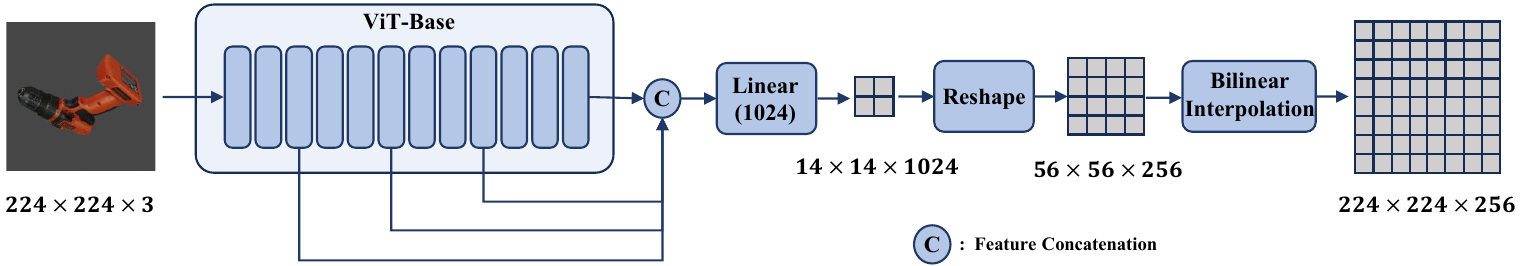}
   \caption{An illustration of the per-pixel feature learning process for an RGB image within the Feature Extraction module of the Pose Estimation Model.}
   \label{fig:pem_feature_arch}
\end{figure*}

In the Pose Estimation Model (PEM) of our SAM-6D, the Feature Extraction module utilizes the base version of the Visual Transformer (ViT) backbone \cite{ViT}, termed as ViT-Base,  to process masked RGB image crops of observed object proposals or rendered object templates, yielding per-pixel feature maps. 

Fig. \ref{fig:pem_feature_arch} gives an illustration of the per-pixel feature learning process for an RGB image within the Feature Extraction module. More specifically, given an RGB image of the object, the initial step involves image processing, including masking the background, cropping the region of interest, and resizing it to a fixed resolution of $224 \times 224$. The object mask and bounding box utilized in the process can be sourced from the Instance Segmentation Model (ISM) for the observed scene image or from the renderer for the object template. The processed image is subsequently fed into ViT-Base to extract per-patch features using 12 attention blocks. The patch features from the third, sixth, ninth, and twelfth blocks are subsequently concatenated and passed through a fully-connected layer. They are then reshaped and bilinearly interpolated to match the input resolution of $224 \times 224$ with 256 feature channels. Further specifics about the network can be found in Fig. \ref{fig:pem_feature_arch}.

For a cropped observed RGB image, the pixel features within the mask are ultimately chosen to correspond to the point set transformed from the masked depth image. For object templates, the pixels within the masks across views are finally aggregated, with the surface point of per pixel known from the renderer. Both point sets of the proposal and the target object are normalized to fit a unit sphere by dividing by the object scale, effectively addressing the variations in object scales.

We use two views of object templates for training, and 42 views for evaluation as CNOS \cite{CNOS}, which is the standard setting for the results reported in this paper.

\subsubsection{Coarse Point Matching}
\label{subsec:supp_pem_arch_2}

In the Coarse Point Matching module, we utilize $T^c$ Geometric Transformers \cite{qin2022geometric} to model the relationships between the sparse point set $\mathcal{P}_m^c \in \mathbb{R}^{N^c_m\times 3}$ of the observed object proposal $m$ and the set $\mathcal{P}_o^c \in \mathbb{R}^{N^c_o\times 3}$ of the target object $\mathcal{O}$. Their respective features $\bm{F}_m^{c}$ and $\bm{F}_o^{c}$ are thus improved to their enhanced versions $\tilde{\bm{F}}_m^{c}$ and $\tilde{\bm{F}}_o^{c}$. Each of these enhanced feature maps also includes the background token. An additional fully-connected layer is applied to the features both before and after the transformers. In this paper, we use the upper script `c' to indicate variables associated with the Coarse Point Matching module, and the lower scripts `m' and `o' to distinguish between the proposal and the object.

During inference, we compute the soft assignment matrix $\tilde{\mathcal{A}}^c \in \mathbb{R}^{(N^c_m+1) \times (N^c_o+1)}$, and obtain two binary-value matrices $\bm{M}_m^c \in \mathcal{R}^{N^c_m\times 1}$ and $\bm{M}_o^c \in \mathcal{R}^{N^c_o \times 1}$, denoting whether the points in $\mathcal{P}_m^c$ and $\mathcal{P}_o^c$ correspond to the background, owing to the design of background tokens; `0' indicates correspondence to the background, while `1' indicates otherwise. We then have the probabilities $\bm{P}^c \in \mathbb{R}^{N^c_m \times N^c_o}$ to indicate the matching degree of the $N^c_m \times N^c_o$ point pairs between $\mathcal{P}_m^c$ and $\mathcal{P}_o^c$, formulated as follows:
\begin{equation}
    \bm{P}^c = \bm{M}_m^c \cdot (\tilde{\mathcal{A}}^c[1:, 1:])^\gamma  \cdot \bm{M}_o^{cT},
\end{equation}
where $\gamma$ is used to sharpen the probabilities and set as $1.5$. The probabilities of points that have no correspondence, whether in $\mathcal{P}_m^c$ or $\mathcal{P}_o^c$,  are all set to 0. Following this, the probabilities $\bm{P}^c$ are normalized to ensure their sum equals 1, and act as weights used to randomly select 6,000 triplets of point pairs from the total pool of $N^c_m \times N^c_o$ pairs. Each triplet, which consists of three point pairs, is utilized to calculate a pose using SVD, along with a distance between the point pairs based on the computed pose. Through this procedure, a total of 6,000 pose hypotheses are generated, and to minimize computational cost, only the 300 poses with the smallest point pair distances are selected. Finally, the initial pose for the Fine Point Matching module is determined from these 300 poses, with the pose that has the highest pose matching score being selected.

In the Coarse Point Matching module, we set $T^c=3$ and $N^c_m = N^c_o = 196$, with all the feature channels designated as 256. The configurations of the Geometric Transformers adhere to those used in \cite{qin2022geometric}.

\subsubsection{Fine Point Matching}
\label{subsec:supp_pem_arch_3}

\begin{figure}[t]
  \centering
   \includegraphics[width=0.98\linewidth]{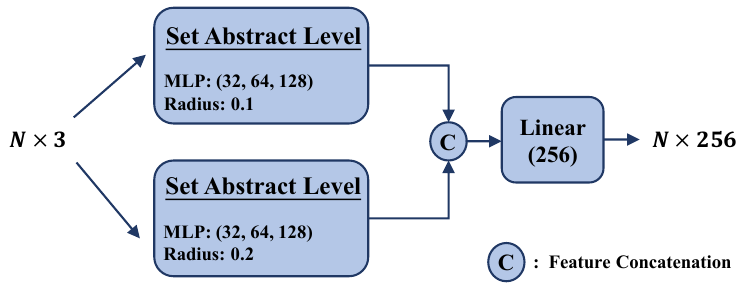}
   \caption{An illustration of the positional encoding for a point set with $N$ points within the Fine Point Matching Module of the Pose Estimation Model. } 
   \label{fig:pem_pos_enc}
\end{figure}

In the Fine Point Matching module, we utilize $T^f$ Sparse-to-Dense Point Transformers to model the relationships between the dense point set $\mathcal{P}_m^f \in \mathbb{R}^{N^f_m\times 3}$ of the observed object proposal $m$ and the set $\mathcal{P}_o^f \in \mathbb{R}^{N^f_o\times 3}$ of the target object $\mathcal{O}$. Their respective features $\bm{F}_m^{f}$ and $\bm{F}_o^{f}$ are thus improved to their enhanced versions $\tilde{\bm{F}}_m^{f}$ and $\tilde{\bm{F}}_o^{f}$. Each of these enhanced feature maps also includes the background token. An additional fully-connected layer is applied to the features both before and after the transformers. We use the upper script `$f$' to indicate variables associated with the Fine Point Matching module, and the lower scripts `m' and `o' to distinguish between the proposal and the object. 

Different from the coarse module, we condition both features $\bm{F}_m^{f}$ and $\bm{F}_o^{f}$ before applying them to the transformers by adding their respective positional encodings, which are learned via a multi-scale Set Abstract Level \cite{qi2017pointnet++} from $\mathcal{P}_m^f$ transformed by the initial pose and $\mathcal{P}_o^f$ without transformation, respectively. The used architecture for positional encoding learning is illustrated in Fig. \ref{fig:pem_pos_enc}. For more details, one can refer to \cite{qi2017pointnet++}. 

\begin{figure*}[t]
  \centering
   \includegraphics[width=0.9\linewidth]{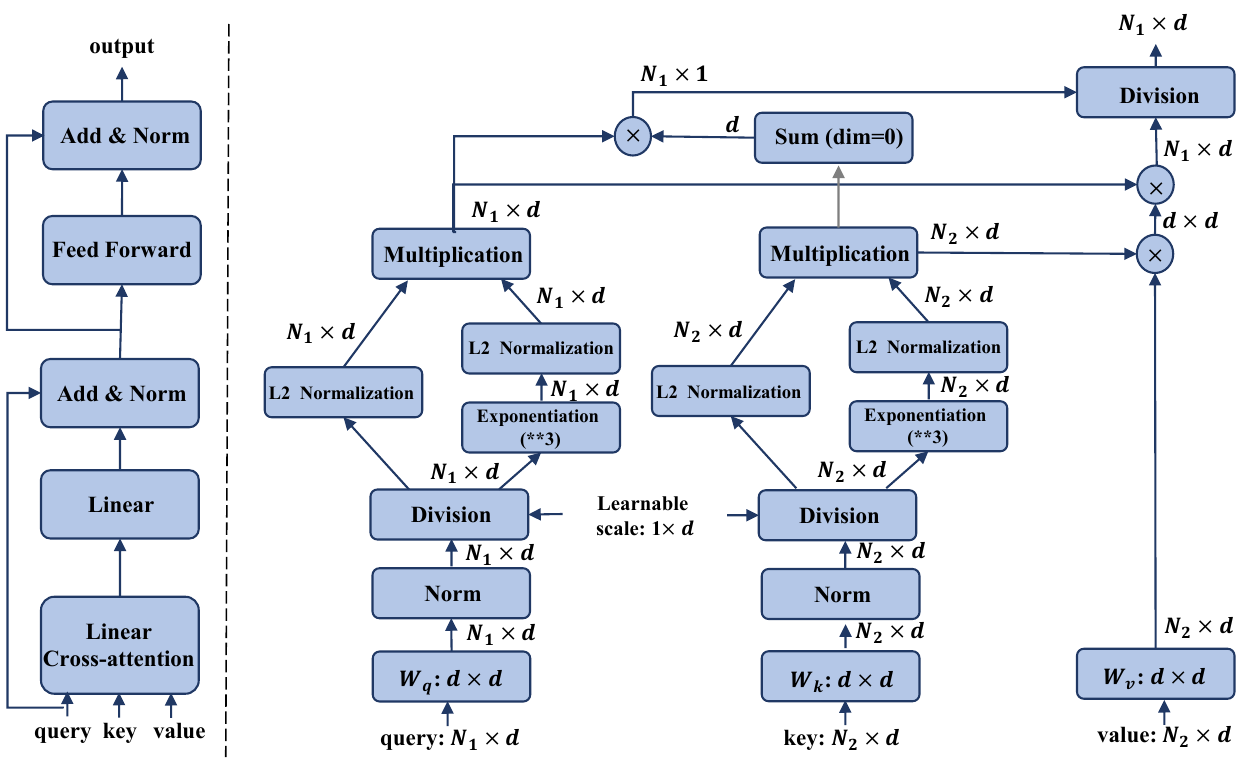}
   \caption{Left: The structure of Linear Cross-attention layer. Right: The structure of Linear Cross-attention.} 
   \label{fig:pem_linear_atten}
\end{figure*}

Another difference from the coarse module is the type of transformers used. To handle dense relationships, we design the Sparse-to-Dense Point Transformers, which utilize Geometric Transformers \cite{qin2022geometric} to process sparse point sets and disseminate information to dense point sets via Linear Cross-attention layers \cite{katharopoulos2020transformers, han2023flatten}. The configurations of the Geometric Transformers adhere to those used in \cite{qin2022geometric}; the point numbers of the sampled sparse point sets are all set as 196. The Linear Cross-attention layer enables attention along the feature dimension, and details of its architecture can be found in Fig. \ref{fig:pem_linear_atten}; for more details, one can refer to \cite{katharopoulos2020transformers, han2023flatten}.

During inference, similar to the coarse module, we compute the soft assignment matrix $\tilde{\mathcal{A}}^f \in \mathbb{R}^{(N^f_m+1) \times (N^f_o+1)}$, and obtain two binary-value matrices $\bm{M}_m^f \in \mathcal{R}^{N^f_m\times 1}$ and $\bm{M}_o^f \in \mathcal{R}^{N^f_o \times 1}$. We then formulate the probabilities $\bm{P}^f \in \mathbb{R}^{N^f_m \times N^f_o}$ as follows:
\begin{equation}
    \bm{P}^f = \bm{M}_m^f \cdot (\tilde{\mathcal{A}}^f[1:, 1:])  \cdot \bm{M}_o^{fT}.
\end{equation}
Based on $\bm{P}^f$, we search for the best-matched point in  $\mathcal{P}_o^f$ for each point in $\mathcal{P}_m^f$, assigned with the matching probability. The final object pose is then calculated using a weighted SVD, with the matching probabilities of the point pairs serving as the weights.

Besides, we set $T^f=3$ and $N^f_m = N^f_o = 2,048$, with all the feature channels designated as 256. During training, we follow \cite{MegaPose} to obtain the initial object poses by augmenting the ground truth ones with random noises.

\subsection{Training Objectives}
\label{subsec:supp_pem_obj}

We use InfoNCE loss \cite{oord2018representation} to supervise the learning of attention matrices for both coarse and fine modules. Specifically, given two point sets $\mathcal{P}_m \in \mathbb{R}^{N_m\times 3}$ and  $\mathcal{P}_o \in \mathbb{R}^{N_o\times 3}$, along with their enhanced features $\tilde{\bm{F}}_m$ and $\tilde{\bm{F}}_o$, which are learnt via the transformers and equipped with background tokens, we compute the attention matrix $\mathcal{A} = \tilde{\bm{F}}_m \times \tilde{\bm{F}}_o^T \in \mathbb{R}^{(N_m+1) \times (N_o+1)}$. Then $\mathcal{A}$ can be supervised by the following objective:
\begin{equation}
    \mathcal{L} = \texttt{CE}(\mathcal{A}[1:, :], \hat{\mathcal{Y}}_m) + \texttt{CE}(\mathcal{A}[:, 1:]^T, \hat{\mathcal{Y}}_o),
    \label{eqn:objective}
\end{equation}
where $\texttt{CE}(\cdot, \cdot)$ denotes the cross-entropy loss function. $\hat{\mathcal{Y}}_m \in \mathbb{R}^{N_m}$ and $\hat{\mathcal{Y}}_o \in \mathbb{R}^{N_o}$ denote the ground truths for $\mathcal{P}_m$ and $\mathcal{P}_o$. Given the ground truth pose $\hat{\bm{R}}$ and $\hat{\bm{t}}$, each element $y_m$ in $\hat{\mathcal{Y}}_m$, corresponding to the point $\bm{p}_m$ in $\mathcal{P}_m$, could be obtained as follows:
\begin{equation}
    y_m =\left\{\begin{matrix}
  0 & \text{if}\enspace d_{k^*} \ge \delta_{dis}  \\
  k^* & \text{if}\enspace d_{k^*} < \delta_{dis}
\end{matrix}\right.,
\end{equation}
where 
\begin{equation}
k^* = \text{Argmin}_{k=1,\dots, N_m} || \hat{\bm{R}}(\bm{p}_m -\hat{\bm{t}})-\bm{p}_{o,k} ||_2, \notag
\end{equation}
and
\begin{equation}
d_{k^*} = || \hat{\bm{R}}(\bm{p}_m -\hat{\bm{t}})-\bm{p}_{o,_{k^*}} ||_2. \notag
\end{equation}
$k^*$ is the index of the closest point $\bm{p}_{o,_{k^*}}$ in $\mathcal{P}_o$ to $\bm{p}_m$, while $d_{k^*}$ denotes the distance between $\bm{p}_m$ and  $\bm{p}_{o,_{k^*}}$ in the object coordinate system. $\delta_{dis}$ is a distance threshold determining whether the point $\bm{p}_m$ has the correspondence in $\mathcal{P}_o$; we set $\delta_{dis}$  as a constant $0.15$,  since both $\mathcal{P}_m$ and $\mathcal{P}_o$ are normalized to a unit sphere. The elements in $\hat{\mathcal{Y}}_o$ are also generated in a similar way.

We employ the objective (\ref{eqn:objective}) upon all the transformer blocks of both coarse and fine point matching modules, and thus optimize the Pose Estimation Model by solving the following problem:
\begin{equation}
    \min \sum_{l=1,\dots, T_c} \mathcal{L}_l^c + \sum_{l=1,\dots, T_f} \mathcal{L}_l^f.
\end{equation}
where for the loss $\mathcal{L}$ in Eq. (\ref{eqn:objective}),  we use the upper scripts `$c$' and `$f$' to distinguish between the losses in the coarse and fine point matching modules, respectively, while the lower script '$l$' denotes the sequence of the transformer blocks in each module.

\subsection{More Quantitative Results}
\label{subsec:supp_pem_quan}

\subsubsection{Effects of The View Number of Templates}
\label{subsec:supp_pem_quan_view}

We present a comparison of results using different views of object templates in Table \ref{tab:pose_nview}. As shown in the table, results with only one template perform poorly as a single view cannot fully depict the entire object. With an increase in the number of views, performance improves. For consistency with our Instance Segmentation Model and CNOS \cite{CNOS}, we utilize 42 views of templates as the default setting in the main paper.

\begin{table}[h]
  \centering
  \begin{tabular}{c|ccccc}  
    \toprule
    $\#$ View & 1 & 2 & 8 & 16 & 42 \\
    \midrule
    AR & 21.8 & 62.7 & 83.9 & 84.1 & 84.5\\
    \bottomrule
  \end{tabular}
  \caption{Pose estimation results with different view numbers of object templates on YCB-V. We report the mean Average Recall (AR) among VSD, MSSD and MSPD.}
  \label{tab:pose_nview}
\end{table}

\subsubsection{Comparisons with OVE6D}
\label{subsec:supp_pem_quan_ove6d}
OVE6D \cite{cai2022ove6d} is a classical method for zero-shot pose estimation based on image matching, which first constructs a codebook from the object templates for viewpoint rotation retrieval and subsequently regresses the in-plane rotation. When comparing our SAM-6D with OVE6D using their provided segmentation masks (as shown in Table \ref{tab:pose_ove6d}), SAM-6D outperforms OVE6D on LM-O dataset, without the need for using Iterative Closest Point (ICP) algorithm for post-optimization.

\begin{table}[h]
    \centering
    \begin{tabular}{@{}l|c}
      \toprule
      Method &  LM-O\\
      \midrule
      OVE6D \cite{cai2022ove6d}  & 56.1 \\
      OVE6D with ICP \cite{cai2022ove6d}  & 72.8  \\
      SAM-6D (Ours)  & \textbf{74.7} \\
      \bottomrule
    \end{tabular}
    \caption{Quantitative results of OVE6D \cite{cai2022ove6d} and our SAM-6D on LM-O dataset. The evaluation metric is the standard ADD(-S) for pose estimation. SAM-6D is evaluated with the same masks provided by \cite{cai2022ove6d}.}
    \label{tab:pose_ove6d}
\end{table}

\begin{figure*}[t]
  \centering
   \includegraphics[width=1.0\linewidth]{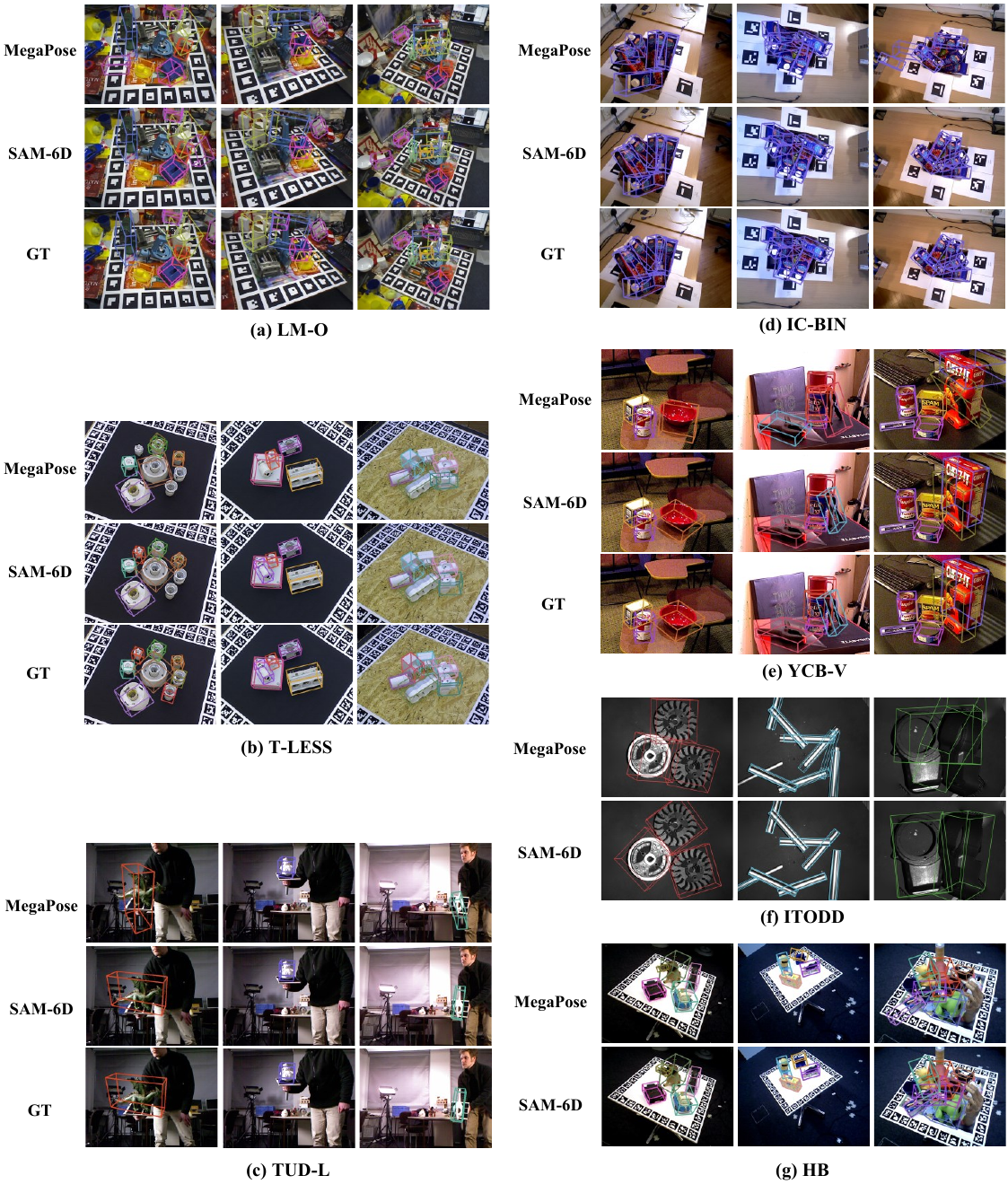}
   \caption{Qualitative results on the seven core datasets of the BOP benchmark \cite{BOP} for pose estimation of novel objects.}
   \label{fig:pose_7dataset}
\end{figure*}

\subsection{More Qualitative Comparisons with Existing Methods}
\label{subsec:supp_pem_qual}
To illustrate the advantages of our Pose Estimation Model (ISM), we visualize in Fig. \ref{fig:pose_7dataset} the qualitative comparisons with MegaPose \cite{MegaPose} on all the seven core datasets of the BOP benchmark \cite{BOP} for pose estimation of novel objects. For reference, we also present the corresponding ground truths, barring those for the ITODD and HB datasets, as these are unavailable.

\end{document}